\documentclass[journal]{IEEEtran}
\usepackage[comma,numbers,square,sort&compress]{natbib}
\usepackage{subfigure}
\usepackage{color}
\usepackage{colortbl}
\usepackage{graphicx}
\usepackage{float}
\usepackage{amsmath,latexsym,amssymb,amsthm,array,amsfonts,algorithm,algpseudocode,booktabs,graphicx,subfigure,multirow,cuted,stfloats}
\usepackage[sort&compress]{natbib}
\usepackage{url}
\usepackage{multirow}
\usepackage{lscape}
\usepackage{nomencl}
\makenomenclature
\usepackage{makecell}
\usepackage{slashbox}
\usepackage{float}
\usepackage{enumerate}
\usepackage{bm}
\definecolor{mygray}{gray}{0.8}
\theoremstyle{definition}

\usepackage{array}

\RequirePackage{ifthen}
\renewcommand{\nomgroup}[1]{%
\ifthenelse{\equal{#1}{A}}{\item[\emph{\textbf{Production line-related notations}}]}{%
\ifthenelse{\equal{#1}{B}}{\item[\emph{\textbf{Production order-related notations}}]}{
\ifthenelse{\equal{#1}{C}}{\item[\emph{\textbf{Other notations}}]}{}}}}

\newcommand{\tabincell}[2]{\begin{tabular}{@{}#1@{}}#2\end{tabular}}
\newlength\savewidth

\newtheorem{remark}{\it Remark}

\begin{document}
\hyphenpenalty=5000
\tolerance=1200
\title{Robust Order Scheduling in the Fashion Industry: A Multi-Objective Optimization Approach}

\author{Wei Du,
        Yang Tang,
        Sunney Yung Sun Leung,
        Le Tong, \\
        Athanasios V. Vasilakos,
        and Feng Qian\\

\thanks{W. Du, Y. Tang and F. Qian are with the Key Laboratory of Advanced Control and Optimization for Chemical Processes, Ministry of Education, East China University of Science and Technology, Shanghai 200237, China (e-mail: duwei0203@gmail.com; tangtany@gmail.com, yangtang@ecust.edu.cn; fqian@ecust.edu.cn).}

\thanks{S. Y. S. Leung and L. Tong are with the Institute of Textile and Clothing, The Hong Kong Polytechnic University, Hong Kong, China (e-mail: sunney.leung@polyu.edu.hk; tongle0328@gmail.com).}

\thanks{A. V. Vasilakos is with the Department of Computer Science, Electrical and Space Engineering, Lulea University of Technology, Lulea 97187, Sweden (e-mail: vasilako@ath.forthnet.gr).}}

\markboth{Preprint submitted to arXiv}%
{Shell \MakeLowercase{\textit{et al.}}: Bare Demo of IEEEtran.cls
for Journals}

\maketitle

\begin{abstract}
In the fashion industry, order scheduling focuses on the assignment of production orders to appropriate production lines. In reality, before a new order can be put into production, a series of activities known as pre-production events need to be completed. In addition, in real production process, owing to various uncertainties, the daily production quantity of each order is not always as expected. In this research, by considering the pre-production events and the uncertainties in the daily production quantity, robust order scheduling problems in the fashion industry are investigated with the aid of a multi-objective evolutionary algorithm (MOEA) called nondominated sorting adaptive differential evolution (NSJADE).
The experimental results illustrate that it is of paramount importance to consider pre-production events in order scheduling problems in the fashion industry. We also unveil that the existence of the uncertainties in the daily production quantity heavily affects the order scheduling.
%\boldmath
\end{abstract}


% \fi
\IEEEpeerreviewmaketitle

\section{Introduction}
Order scheduling is a key decision-making problem in supply chain management of manufacturing industry, and plays an important role in rational resource allocation and utilization, which makes the companies more competitive in the global market \cite{chan2006early,brankeautomated}. As a typical representative of labor-intensive industries, the fashion industry is characterized by short product life cycles, volatile customer demands, rising labor costs, tremendous product varieties, and long supply processes \cite{csen2008us}. Therefore, more attention needs to be paid to fashion order scheduling problems.

For the past few decades, order scheduling problems in the fashion industry have been widely investigated \cite{chen2006order,guo2013modeling,wong2014intelligent}.
In these researches, it is assumed that all the orders are ready for production when the production process begins. However, in real-world apparel production, a whole series of activities need to be carried out before an order can be put into production. And these activities are known as pre-production events. In the fashion industry, pre-production events include order fabric, order trims, sample approval, issue markers, and so on \cite{hines2007fashion}. Multiple parties, e.g., suppliers, manufacturers, and customers, need to collaborate with each other to complete a pre-production event. And negotiations among them are time-consuming and uncertain, which results in the late completion of some pre-production events and the delay of producing the related order \cite{preprod}.
In addition, industrial data concerning the pre-production process are difficult to collect. Therefore, pre-production events have been largely overlooked in the order scheduling research of the fashion industry, which makes the first incentive of this paper.

In recent years, as a powerful optimization tool \cite{eiben2015evolutionary,du2017differential}, evolutionary algorithms (EAs) have been introduced to solve the order scheduling problems in the fashion industry \cite{wong2014intelligent,guo2013hybrid}. In the studies above, when the schedules were made before the real production, it was assumed that the daily production quantity of each order was fixed. However, in most real-world manufacturing environments of the fashion industry, order scheduling is an ongoing reactive process in which the occurrence of various unexpected disruptions are usually inevitable \cite{ouelhadj2009survey}. These disruptions consist of machine breakdown, operator absenteeism, and so on. Therefore, the daily production quantity of each order is not always as expected in the production process. As a result, the pre-established order schedules are shifted very often after the production starts. However, a frequent modification of order schedules will increase labor and time cost, which may reduce production efficiency and fail to complete the orders before their delivery dates. Therefore, the second incentive of this paper originates from considering robust order schedules, which are not sensitive to the variation of the daily production quantity during the process of the real production.

For the past decade and more, robust optimization has gained increasing attention, and has been incorporated into the framework of single-objective evolutionary optimization \cite{jin2003trade,jin2005evolutionary}. In the context of multi-objective optimization, Deb and Gupta \cite{deb2006introducing} did some pioneering works by suggesting two different ways of introducing robustness in multi-objective optimization. Up to now, the integration of robust multi-objective optimization and MOEA has been gradually applied to deal with a variety of applications, such as the welded beam design problem \cite{deb2006introducing} and the controllability of complex neuronal networks \cite{tang2015robust}. Therefore, in searching for a candidate for robust order schedules in the fashion industry, robust MOEA can be selected as a promising one.

Based on the above discussion, in this paper, robust order scheduling is presented via introducing robust multi-objective optimization into order scheduling problems in the fashion industry. In addition, the pre-production events in apparel manufacturing are also taken into account and the order scheduling problem is modelled as a multi-objective optimization problem. A MOEA called nondominated sorting adaptive differential evolution (NSJADE) is utilized to search the order schedules in the fashion industry that achieve the following three objectives: 1) the schedules can minimize the total pre-production event clashes of all orders; 2) the schedules can minimize the total tardiness of all orders; 3) the schedules are not sensitive to variation of the daily production quantity during the process of real production. The contributions of this paper are mainly threefold: 1) to the best of our knowledge, it is the first attempt in which the pre-production events are considered for the order scheduling research in the fashion industry; 2) robust order schedules are obtained with the aid of robust multi-objective optimization combined with NSJADE; 3) compared with the results obtained by adaptive differential evolution (JADE) and NSJADE without uncertainty, it is revealed that the pre-production events and the existence of the uncertainties in the daily production quantity heavily affect the order scheduling.

\section{Problem Description and Formulation}\label{sec1}
In this section, the problem of robust order scheduling is described in detail. The settings of the notations and the variables considered in the problem are on the basis of a business software called Fast React \cite{fastreact}, which is specifically for the fashion industry.

\nomenclature[Ca]{$\emph{S}_\emph{day}$}{the day when making the order schedule}%
\nomenclature[Cb]{$\emph{P}_\emph{day}$}{the day when the production begins}%
\nomenclature[Ac]{$\emph{m}$}{the number of production lines}%
\nomenclature[Ad]{$\emph{P}_\emph{i}$}{the \emph{i}th production line ($1\leq\emph{i}\leq\emph{m}$)}%
\nomenclature[Be]{$\emph{n}$}{the number of production orders}%
\nomenclature[Bf]{$\emph{O}_\emph{j}$}{the \emph{j}th production order ($1\leq\emph{j}\leq\emph{n}$)}%
\nomenclature[Cg]{$\emph{p}$}{the number of product types}%
\nomenclature[Ch]{$\emph{T}_\emph{l}$}{the \emph{l}th product type ($1\leq\emph{l}\leq\emph{p}$)}%
\nomenclature[Ai]{$\emph{E}_\emph{p,il}$}{efficiency of production line $\emph{P}_\emph{i}$ for producing order of type $\emph{T}_\emph{l}$}%
\nomenclature[Aj]{$\emph{C}^\emph{mins}_\emph{i}$}{capacity minutes per day of production line $\emph{P}_\emph{i}$}%
\nomenclature[Bk]{$\emph{R}_\emph{j}$}{product type of order $\emph{O}_\emph{j}$}%
\nomenclature[Bl]{$\emph{Q}_\emph{j}$}{quantity of order $\emph{O}_\emph{j}$}%
\nomenclature[Bm]{$\emph{S}^\emph{mins}_\emph{j}$}{standard minutes per piece for order $\emph{O}_\emph{j}$}%
\nomenclature[Bn]{$\emph{A}_\emph{day,j}$}{scheduled starting date of order $\emph{O}_\emph{j}$}%
\nomenclature[Bo]{$\emph{F}_\emph{day,j}$}{scheduled finishing date of order $\emph{O}_\emph{j}$}%
\nomenclature[Bp]{$\emph{D}_\emph{day,j}$}{due date of order $\emph{O}_\emph{j}$}%
\nomenclature[Bq]{$\emph{C}_\emph{day,j}$}{present conservative starting date of order $\emph{O}_\emph{j}$}%
\nomenclature[Br]{$\emph{N}_\emph{j}$}{the number of the pre-production events of order $\emph{O}_\emph{j}$}%
\nomenclature[Bs]{$\emph{F}_\emph{jk}$}{offset days of the \emph{k}th pre-production event of order $\emph{O}_\emph{j}$ ($1\leq\emph{k}\leq\emph{N}_\emph{j}$)}%
\nomenclature[Bt]{$\emph{X}_\emph{jk}$}{indicates if the \emph{k}th pre-production event of order $\emph{O}_\emph{j}$ is finished on day $\emph{S}_\emph{day}$, $\emph{X}_\emph{jk}=1$; otherwise, $\emph{X}_\emph{jk}=0$ ($1\leq\emph{k}\leq\emph{N}_\emph{j}$).}%
\nomenclature[Cu]{$\emph{f}_\emph{l}(\cdot)$}{a function indicating the learning curve of product type $\emph{T}_\emph{l}$}%
\nomenclature[Bu1]{$\emph{E}_\emph{o,j}$}{efficiency of producing order $\emph{O}_\emph{j}$}%
\nomenclature[Cu2]{$\emph{U}_\emph{l}$}{consecutive days of producing order of type $\emph{T}_\emph{l}$}%
\nomenclature[Bu3]{$\emph{P}_\emph{time,j}$}{production time of order $\emph{O}_\emph{j}$}%
\nomenclature[Bu4]{$\emph{N}_\emph{time,j}$}{time needed to reach the next efficiency level for order $\emph{O}_\emph{j}$}%
\nomenclature[Bu5]{$\emph{S}_\emph{time,j}$}{time spent on order $\emph{O}_\emph{j}$ on the current efficiency level}%
\nomenclature[Bu6]{$\emph{Q}_\emph{jd}$}{quantity of order $\emph{O}_\emph{j}$ completed on the \emph{d}th day of processing $\emph{O}_\emph{j}$ ($1\leq\emph{d}\leq\emph{F}_\emph{day,j}-\emph{A}_\emph{day,j}$)}%
\nomenclature[Bu7]{$\emph{Q}_\emph{sum,jd}$}{total quantity of order $\emph{O}_\emph{j}$ completed from the first day till the \emph{d}th day of processing $\emph{O}_\emph{j}$ ($1\leq\emph{d}\leq\emph{F}_\emph{day,j}-\emph{A}_\emph{day,j}$)}%
\nomenclature[Bv]{$\emph{q}$}{the number of sub-orders of order $\emph{O}_\emph{j}$}%
\nomenclature[Bw]{$\emph{O}_\emph{jr}$}{the \emph{r}th sub-order of order $\emph{O}_\emph{j}$ ($1\leq\emph{r}\leq\emph{q}$)}%
\nomenclature[Bx]{$\alpha_\emph{j}$}{split percentage of order $\emph{O}_\emph{j}$}%
\nomenclature[Cy]{$\beta$}{uncertainty factor of daily production quantity}%
\renewcommand{\nomname}{\normalsize{N}\footnotesize{OMENCLATURE}}
\printnomenclature

\subsection{Problem Description} \label{odattri}
\begin{figure}[!htbp]
\begin{minipage}[t]{1\linewidth}
\centering
\includegraphics[width=8.8cm]{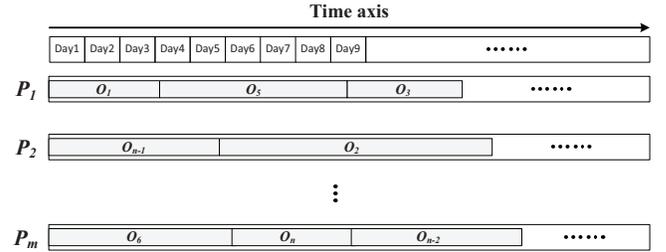}
\caption{An illustration of the robust order scheduling problem in the fashion industry.} \label{illustr}
\end{minipage}
\end{figure}

The robust order scheduling problem in the fashion industry considers \emph{m} production lines and \emph{n} production orders, and \emph{n} orders are assigned to appropriate lines for production. An illustration is displayed in Fig. \ref{illustr}. In Fig. \ref{illustr}, the order bar represents the duration of producing the related order.
\subsubsection{Description of Production Line}
The production lines consist of a number of different machines so that a garment can start at the top of the line in its cut state and come off the line once completed. On a production line, the machines are responsible for cutting, embroidery, printing, sewing, pressing and packing, respectively; and sewing is the most time-consuming process. Since we focus on the impact of pre-production events and the uncertainties in the daily production quantity on the fashion order scheduling, to simplify the problem, we only consider the sewing process during the production in this paper, instead of all the processes. In this research, production lines are product-specific lines, which means that the line's efficiency is lower than its peak when there is a mismatch of product to production line.

\begin{table}[!htbp]
\footnotesize
\centering
\caption{An example of the description of 2 production lines. Product type of blouses is marked as $\emph{T}_\emph{1}$, product type of jackets is marked as $\emph{T}_\emph{2}$.}\label{egpl}
{\begin{tabular}{|c|c|c|c|}
\hline
\tabincell{c}{\textbf{Production} \\ \textbf{Line}} & \tabincell{c}{\textbf{Efficiency for} \\ \textbf{Blouses (\%)}} & \tabincell{c}{\textbf{Efficiency for} \\ \textbf{Jackets (\%)}} & \tabincell{c}{\textbf{Capacity} \\ \textbf{(mins/day)}} \\\hline
$\emph{P}_\emph{1}$  & 100 & 80 & 6720  \\\hline
$\emph{P}_\emph{2}$  & 80 & 100 & 6240  \\\hline
\end{tabular}}
\end{table}
\begin{table*}[!htbp]
\footnotesize
\centering
\caption{An example of the description of 2 production orders.}\label{egpo}
{\begin{tabular}{|c|c|c|c|c|c|}
\hline
\tabincell{c}{\textbf{Production} \\ \textbf{Order}} & \tabincell{c}{\textbf{Product} \\ \textbf{Type}} & \textbf{Quantity} & \tabincell{c}{\textbf{Present Conservative} \\ \textbf{Starting Date (when} $\emph{\textbf{S}}_\emph{\textbf{day}}\textbf{=-14}$)} & \textbf{Due Date} & \tabincell{c}{\textbf{Standard Minutes} \\ \textbf{Per Piece}} \\\hline
$\emph{O}_\emph{1}$  & Skirts & 870 & 6 & 10 & 14.20  \\\hline
$\emph{O}_\emph{2}$  & Blouses & 800 & 0 & 11 & 18.20  \\\hline
\end{tabular}}
\end{table*}
An example is provided in Table \ref{egpl} to better illustrate the production lines. In Table \ref{egpl}, the second and third columns show that $\emph{P}_\emph{1}$ is blouse-specific line and $\emph{P}_\emph{2}$ is jacket-specific line; if a mismatch of product to production line occurs, the efficiency $\emph{E}_\emph{p,12}$ and $\emph{E}_\emph{p,21}$ will reduce to 80\%. The last column gives the value of ``\emph{capacity minutes per day}" of each line: $\emph{C}^\emph{mins}_\emph{1}=6720$ and $\emph{C}^\emph{mins}_\emph{2}=6240$.

\subsubsection{Description of Production Order}
Each order has five attributes, and an example of 2 production orders is given in Table \ref{egpo}. For convenience, the day when the production begins $\emph{P}_\emph{day}$ is set as day 0 in this paper. ``\emph{Present conservative starting date}" denotes the earliest safe starting date of this order's production, which is determined by its pre-production events which have not been finished on the day $\emph{S}_\emph{day}$. More details of this attribute will be explained in Section \ref{cdpre}. ``\emph{Due date}" shows when the order needs to be completed and delivered to customers. The attribute ``\emph{standard minutes per piece}" is used to represent the workload of the sewing process of each order. In this research, orders can be split into \emph{q} sub-orders, with the purpose of realizing flexible production; the split percentage is denoted by $\alpha_\emph{j}=[\alpha_{\emph{j}1}, \alpha_{\emph{j}2},..., \alpha_{\emph{j}(\emph{q}-1)}]^T$.

\subsubsection{Description of Other Notations}
\begin{figure}[!htbp]
\begin{minipage}[t]{1\linewidth}
\centering
\includegraphics[width=6.5cm]{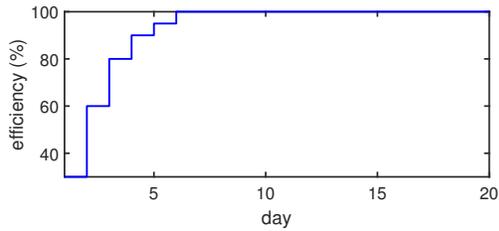}
\caption{An illustration of the learning curve when producing the order of type $\emph{T}_\emph{l}$.} \label{lcplot}
\end{minipage}
\end{figure}

When a new type of product is introduced into production, it takes a period of time for the operators on the production line to get familiar with the production of this type of product. As time goes on, the efficiency of the production line improves day by day as the operators become more familiar with the product and any new manufacturing techniques or skills required. The increase of the efficiency during this period of time is illustrated by ``\emph{learning curve}" ($\emph{f}_\emph{l}(\cdot)$), which is specific according to different types of the product (see Fig. \ref{lcplot}). The uncertainties in the daily production quantity are taken into account in this paper, and represented by $\beta$.

In the following, the problem formulation is separated into two parts, in which the critical variables concerning pre-production events and the uncertainties in the daily production quantity are explained in detail, respectively.

\subsection{Variables Regarding Pre-Production Events}\label{cdpre}
In the robust order scheduling problems, four variables are related to the launch and the termination of order $\emph{O}_\emph{j}$'s production. They are present conservative starting date of order $\emph{O}_\emph{j}$ ($\emph{C}_\emph{day,j}$), due date of order $\emph{O}_\emph{j}$ ($\emph{D}_\emph{day,j}$), scheduled starting date of order $\emph{O}_\emph{j}$ ($\emph{A}_\emph{day,j}$), and scheduled finishing date of order $\emph{O}_\emph{j}$ ($\emph{F}_\emph{day,j}$). Among them, $\emph{D}_\emph{day,j}$ is one of the attributes of order $\emph{O}_\emph{j}$, and $\emph{A}_\emph{day,j}$ and $\emph{F}_\emph{day,j}$ of $\emph{O}_\emph{j}$ can be obtained when the schedule is made. For $\emph{C}_\emph{day,j}$, the value is determined by $\emph{O}_\emph{j}$'s pre-production events which have not been finished on the day $\emph{S}_\emph{day}$. In the following, the process of calculating $\emph{C}_\emph{day,j}$ is introduced in detail.
\begin{table}[!htbp]
\footnotesize
\centering
\caption{Timetable of the pre-production events and the updated progress of the pre-production events of order $\emph{O}_\emph{j}$ when $\emph{S}_\emph{day}=-14$. The unfinished pre-production events are highlighted in grey background.}\label{pre1}
{\begin{tabular}{m{0pt}p{65pt}p{20pt}||p{65pt}p{20pt}}
\hline
\rule{0pt}{8pt}& \textbf{Event Name} & \textbf{Offset Days} & \textbf{Event Name} & \textbf{Offset Days}\\\hline
\rule{0pt}{8pt}& PO Receive & -60 & PO Receive & -60 \\
\rule{0pt}{8pt}& Order Fabric & -55 & Order Fabric & -55 \\
\rule{0pt}{8pt}& Order Trims & -40 & Order Trims & -40 \\
\rule{0pt}{8pt}& Lab Dip Submit & -35 & Lab Dip Submit & -35 \\
\rule{0pt}{8pt}& Lab Dip Approval & -20 & Lab Dip Approval & -20 \\
\rule{0pt}{8pt}& Sample Approval & -15 & \multicolumn{1}{>{\columncolor{mygray}}l}{Sample Approval} & \multicolumn{1}{>{\columncolor{mygray}}l}{-15} \\
\rule{0pt}{8pt}& Fabric Receipt & -10 & \multicolumn{1}{>{\columncolor{mygray}}l}{Fabric Receipt} & \multicolumn{1}{>{\columncolor{mygray}}l}{-10} \\
\rule{0pt}{8pt}& Issue Markers & -7 & \multicolumn{1}{>{\columncolor{mygray}}l}{Issue Markers} & \multicolumn{1}{>{\columncolor{mygray}}l}{-7} \\
\rule{0pt}{8pt}& Trims Receipt & -7 & Trims Receipt & -7 \\
\rule{0pt}{8pt}& Factory PP Meeting & -7 & \multicolumn{1}{>{\columncolor{mygray}}l}{Factory PP Meeting} & \multicolumn{1}{>{\columncolor{mygray}}l}{-7} \\
\hline
\end{tabular}}
\end{table}

For each order, a whole series of activities need to be accomplished before the order can be put into production, and these events are called pre-production events. According to pre-production events, we can set up a timetable, which contains the name of events and the number of working days before the start by which each event needs to be finished. An illustration is displayed in the left half of Table \ref{pre1}. Some of the events are closely linked, like ``Lab Dip Submit" and ``Lab Dip Approval", and the only requirement to begin ``Lab Dip Approval" is that ``Lab Dip Submit" must be finished.

As it gets closer to the start of the production, the progress of the pre-production events will be updated. The right half of Table \ref{pre1} shows the updated progress of the pre-production events of order $\emph{O}_\emph{j}$ when the schedule is made 14 days before the production, i.e., $\emph{S}_\emph{day}=-14$.
The unfinished pre-production events of $\emph{O}_\emph{j}$ are highlighted in grey background. And the earliest safe starting date of producing $\emph{O}_\emph{j}$ is determined by the unfinished pre-production event with the largest offset days, e.g., ``Sample Approval" in Table \ref{pre1}. In this case, the earliest safe starting date of producing $\emph{O}_\emph{j}$ should be $-14+|-15|=1$, since $\emph{S}_\emph{day}=-14$. The method of calculating $\emph{C}_\emph{day,j}$ is concluded below.

For order $\emph{O}_\emph{j}$, assume that $\tilde{\emph{k}}$th pre-production event is not completed and has the largest value of $|\emph{F}_\emph{jk}|$, that is, $|\emph{F}_{\emph{j}\tilde{\emph{k}}}|=\max(|\emph{F}_{\emph{jk}}|\cdot\emph{X}_{\emph{jk}})$, where $1\leq\emph{k}\leq\emph{N}_\emph{j}$. Therefore, order $\emph{O}_\emph{j}$'s present conservative starting date $\emph{C}_\emph{day,j}$ is calculated as follows:
\begin{equation}
 \emph{C}_\emph{day,j}=\left\{
\begin{array}{ll}
\emph{P}_\emph{day}+|\emph{F}_{\emph{j}\tilde{\emph{k}}}|-(\emph{P}_\emph{day}-\emph{S}_\emph{day}), & \textrm{if} \hspace{0.1cm} |\emph{F}_{\emph{j}\tilde{\emph{k}}}|>\emph{P}_\emph{day}-\emph{S}_\emph{day},\\
\emph{P}_\emph{day},       & \textrm{otherwise}.
\end{array} \right.
\label{eq10}
\end{equation}

To sum up, the introduction of pre-production events influences the order's present conservative starting date, which mainly depends on the order's pre-production events which have not been finished on the day when making the order schedule.

\begin{remark}
In this research, pre-production events are integrated in the scheduling as one of the objective functions. We didn't consider pre-production events as individual tasks, which was the research paradigm mentioned in other preparation action-related works \cite{berger2004many,sawik2014joint}. The reasons can be summarized as follows: 1) In the manufacturing environment of the fashion industry, production is closely related to pre-production events, and an order cannot be put into production until its pre-production events are all completed; 2) In a schedule, pre-production events influence the starting date of an order, while production determines the finishing date of this order.
In addition, the negotiations among multiple parties to fulfill an order's pre-production events are time-consuming and uncertain, which means some events cannot be completed as originally planned. Therefore, a dynamic adjustment of the schedules can be made in terms of the real-time updating information of pre-production events, which is a prominent advantage of integrating pre-production events in the order scheduling. Furthermore, in Section \ref{secc}, we will discuss the difference between integrating and not integrating pre-production events in the scheduling. The experimental results will show that pre-production events are closely linked with the production, and have an important impact on the scheduling problems in the fashion industry.
\end{remark}

\subsection{Variables Regarding the Uncertainties in the Daily Production Quantity}
In a schedule, when order $\emph{O}_\emph{j}$ is assigned to production line $\emph{P}_\emph{i}$, $\emph{O}_\emph{j}$'s scheduled finishing date $\emph{F}_\emph{day,j}$ can be computed in terms of its scheduled starting date $\emph{A}_\emph{day,j}$ and its production time $\emph{P}_\emph{time,j}$. In this research, the uncertainties in the daily production quantity are taken into consideration, hence the calculation of $\emph{P}_\emph{time,j}$ is related to $\beta$, which is the uncertainty factor of daily production quantity.

$\emph{F}_\emph{day,j}$ indicates the date when the production of order $\emph{O}_\emph{j}$ is completed, which is planned in the order schedule made on the day $\emph{S}_\emph{day}$. If order $\emph{O}_\emph{j}$ is not split into sub-orders during the production, $\emph{F}_\emph{day,j}$ is the date when the production of $\emph{O}_\emph{j}$ ends:
\begin{equation}
\emph{F}_\emph{day,j}=\emph{A}_\emph{day,j}+\emph{P}_\emph{time,j}.
\label{eqfd1}
\end{equation}
If order $\emph{O}_\emph{j}$ is split into \emph{q} sub-orders, $\emph{F}_\emph{day,j}$ represents the ending date of producing sub-order $\emph{O}_{\emph{j}\check{\emph{r}}}$ ($1\leq\check{\emph{r}}\leq\emph{q}$) in the schedule, where $\emph{O}_{\emph{j}\check{\emph{r}}}$ is the last sub-order to be finished among all the \emph{q} sub-orders:
\begin{equation}
\emph{F}_\emph{day,j}=\emph{A}_{\emph{day,j}\check{\emph{r}}}+\emph{P}_{\emph{time,j}\check{\emph{r}}}.
\label{eqfd2}
\end{equation}

In the following, we will explain how to calculate the production time of an order. The calculation process is the same no matter whether the order is split. Therefore, we take order $\emph{O}_\emph{j}$ as an example, which is not split during the production. Assume that order $\emph{O}_\emph{j}$ is of type $\emph{T}_\emph{l}$ and processed on production line $\emph{P}_\emph{i}$, and here we have the procedure to calculate $\emph{P}_\emph{time,j}$.

First, determine the efficiency of producing order $\emph{O}_\emph{j}$: $\emph{E}_\emph{o,j}$ and the time needed to reach the next efficiency level for order $\emph{O}_\emph{j}$: $\emph{N}_\emph{time,j}$.
There are two circumstances:

1) If $\emph{O}_\emph{j}$ is not the first order processed on production line $\emph{P}_\emph{i}$, we assume that the order processed right before $\emph{O}_\emph{j}$ is order $\emph{O}_{\hat{\emph{j}}}$ ($1\leq\hat{\emph{j}}\leq\emph{n}$). If $\emph{O}_\emph{j}$ is of the same type as $\emph{O}_{\hat{\emph{j}}}$, the consecutive days of producing order of type $\emph{T}_\emph{l}$: $\emph{U}_\emph{l}$ can keep accumulating instead of re-initialization. Then $\emph{E}_\emph{o,j}$ can be computed according to $\emph{E}_\emph{o,j}=\emph{f}_\emph{l}(\emph{U}_\emph{l})$. $\emph{N}_\emph{time,j}$ can be obtained by the following equation:
\begin{equation}
\emph{N}_\emph{time,j}=\emph{C}^\emph{mins}_\emph{i}-\emph{S}_{\emph{time,}\hat{\emph{j}}},
\label{eqtn}
\end{equation}
where $\emph{S}_{\emph{time,}\hat{\emph{j}}}$ is the time spent on order $\emph{O}_{\hat{\emph{j}}}$ on its current efficiency level.

2) If $\emph{O}_\emph{j}$ is not the first order processed on production line $\emph{P}_\emph{i}$ and $\emph{O}_\emph{j}$ is of the different type from $\emph{O}_{\hat{\emph{j}}}$, or $\emph{O}_\emph{j}$ is the first order processed on production line $\emph{P}_\emph{i}$, $\emph{U}_\emph{l}$ is re-initialized as 1, and $\emph{E}_\emph{o,j}$ can be obtained according to $\emph{E}_\emph{o,j}=\emph{f}_\emph{l}(\emph{U}_\emph{l})$. $\emph{N}_\emph{time,j}$ is initialized as $\emph{C}^\emph{mins}_\emph{i}$.

Second, after having the values of $\emph{E}_\emph{o,j}$ and $\emph{N}_\emph{time,j}$, we can calculate the value of $\emph{F}_\emph{day,j}$ according to Algorithm 1.
In Algorithm 1, Line 11 shows how to calculate the quantity of $\emph{O}_\emph{j}$ completed on the \emph{d}th day of processing $\emph{O}_\emph{j}$, which reflects the impact of considering the uncertainties in the daily production quantity. The while loop terminates when the total quantity of order $\emph{O}_\emph{j}$ completed from the first day till the \emph{d}th day of processing $\emph{O}_\emph{j}$ reaches the quantity of $\emph{O}_\emph{j}$. Finally, $\emph{P}_\emph{time,j}$ and $\emph{F}_\emph{day,j}$ can be obtained.
\begin{algorithm}[!t]
\scriptsize{
\caption{Calculation\_of\_$\emph{F}_\emph{day,j}$ ()} \algblock{Begin}{End}
{\begin{algorithmic}[1]
\Begin
\State /* $\emph{Q}_\emph{sum,jd}$ is the total quantity of order $\emph{O}_\emph{j}$ completed from the first day till the \emph{d}th day of processing $\emph{O}_\emph{j}$
\State /* $\emph{Q}_\emph{j}$ is the quantity of $\emph{O}_\emph{j}$
\State /* $\emph{E}_\emph{p,il}$ is the efficiency of production line $\emph{P}_\emph{i}$ for producing order of type $\emph{T}_\emph{l}$
\State /* $\emph{S}^\emph{mins}_\emph{j}$ is the standard minutes per piece for order $\emph{O}_\emph{j}$
\State /* $\beta$ is the uncertainty factor of daily production quantity
\State /* $\textrm{rand}(\emph{a},\emph{b})$ uniformly generates a random number belonging to the interval $(\emph{a},\emph{b})$
\State $\emph{d}=1$, $\emph{Q}_\emph{sum,jd}=0$
\While {$\emph{Q}_\emph{sum,jd}<\emph{Q}_\emph{j}$}
\State $\emph{E}_\emph{o,j}=\emph{f}_\emph{l}(\emph{U}_\emph{l})$
\State $\emph{Q}_\emph{jd}=\frac{\emph{N}_\emph{time,j}\cdot\emph{E}_\emph{p,il}\cdot\emph{E}_\emph{o,j}}{\emph{S}^\emph{mins}_\emph{i}}\cdot(1+\textrm{rand}(-\beta,\beta))$
\State $\emph{Q}_\emph{sum,jd}=\emph{Q}_\emph{sum,jd}+\emph{Q}_\emph{jd}$
\State $\emph{d}=\emph{d}+1$, $\emph{U}_\emph{l}=\emph{U}_\emph{l}+1$, $\emph{N}_\emph{time,j}=\emph{C}^\emph{mins}_\emph{i}$
\EndWhile
\State $\emph{P}_\emph{time,j}=\emph{d}-1$
\State $\emph{F}_\emph{day,j}=\emph{A}_\emph{day,j}+\emph{P}_\emph{time,j}$
\End
\end{algorithmic}}}
\end{algorithm}

In a word, the uncertainties in the daily production quantity affect the total processing time of the order, and hence has an impact on the order's scheduled finishing date.

\subsection{Objective Functions}
The investigated order scheduling problem aims at making robust order schedules in the fashion industry, in which the pre-production events and the uncertainties in the daily production quantity are both considered. In a schedule, if an order is assigned to be processed on a day before the present conservative starting date, we say that a pre-production event clash occurs; if the order is scheduled to be completed after its due date, we say that there is tardiness for delivering this order.
Therefore, we have two objectives for the robust order scheduling problems: 1) minimizing the total tardiness of all orders; 2) minimizing the total pre-production event clashes of all orders. In the following, these two objectives are introduced in detail.

The first objective is to minimize the total tardiness of all orders, which is described as follows:
\begin{equation}
f_1=\sum\limits_{\emph{j}=1}^{\emph{n}}h(\emph{DD}_\emph{j}-\emph{FD}_\emph{j}),
\label{eqobj1}
\end{equation}
where $\emph{DD}_\emph{j}$ and $\emph{FD}_\emph{j}$ are due date and scheduled finishing date of order $\emph{O}_\emph{j}$, respectively. $h(\cdot)$ is defined as follows:
\begin{equation}
 h(x)=\left\{
\begin{array}{ll}
0, & \textrm{if} \hspace{0.1cm} x\geq0,\\
-x,       & \textrm{otherwise}.
\end{array} \right.
\label{eqh}
\end{equation}

The second objective is to minimize the total pre-production event clashes of all orders, which is expressed as follows:
\begin{equation}
f_2=\sum\limits_{\emph{j}=1}^{\emph{n}}h(\emph{AD}_\emph{j}-\emph{CD}_\emph{j}),
\label{eqobj2}
\end{equation}
where $\emph{AD}_\emph{j}$ and $\emph{CD}_\emph{j}$ are scheduled starting date and present conservative starting date of order $\emph{O}_\emph{j}$, respectively.
These two objectives determine the assignment of all the orders on the production lines, and they conflict with each other, which means the solution leading to a smaller $f_1$ (less total tardiness) can cause a larger $f_2$ (more total pre-production event clashes). Therefore, the robust order scheduling problem in the fashion industry can be modelled as a multi-objective optimization problem.

\begin{remark}
In apparel manufacturing, the production starts as planned even if some pre-production events of individual orders have not been completed, which is realized by moving the orders with unfinished pre-production events to the later stage of the production, and arranging the orders of which the pre-production events are all finished to be produced at first.
Fashion order scheduling problems depend much on the progress of the orders' pre-production events. It is quite often that some events of individual orders may fail to be finished as planned since the negotiation process is full of uncertainty. On the basis of our experience in the fashion industry, the schedules are made a period of time preceding the production, and we keep modifying these schedules according to the dynamic updating information of the pre-production events as production approaching. It is worth mentioning that the intention of this paper is to provide the planners with early warnings of the orders with unfinished pre-production events. And by means of evolutionary algorithms, the schedules can be promptly and intelligently updated.
\end{remark}

\section{Robust Multi-Objective Optimization for Order Scheduling Problems in the Fashion Industry}
In this section, the concept of robust multi-objective optimization is first provided. Then a nondominated sorting adaptive differential evolution (NSJADE)-based optimization process is proposed to obtain the robust order schedules in the fashion industry.

\subsection{Robust Multi-Objective Optimization}\label{defH}
As stated in \cite{deb2006introducing}, robustness is introduced in multi-objective optimization by means of optimizing the mean effective objective functions instead of optimizing the original objective functions. Hence the robust multi-objective optimization problem can be formulated as follows:

\emph{Problem:} A solution $\textbf{x}^\ast$ is called a multi-objective robust solution, if it is the global feasible Pareto-optimal solution to the following multi-objective minimization problem (defined with respect to a $\delta$-neighborhood of a solution $\textbf{x}$):
\begin{equation}
\textrm{minimize}\hspace{0.3cm} (f_1^{\textrm{eff}}(\textbf{x}), f_2^{\textrm{eff}}(\textbf{x}),..., f_M^{\textrm{eff}}(\textbf{x})), \hspace{0.3cm} \textbf{x}\in\Omega,
\label{eqproblem}
\end{equation}
where $f_i^{\textrm{eff}}(\textbf{x})$ is defined as follows:
\begin{equation}
f_i^{\textrm{eff}}(\textbf{x})=\frac{1}{|\mathcal{B}_\delta(\textbf{x})|}\int_{\textbf{y}\in\mathcal{B}_\delta(\textbf{x})}f_i(\textbf{y})d\textbf{y},
\label{eqprofi}
\end{equation}
$\mathcal{B}_\delta(\textbf{x})$ is a $\delta$-neighborhood of a solution $\textbf{x}$, $|\mathcal{B}_\delta(\textbf{x})|$ is the hypervolume of the neighborhood; $\Omega$ is the feasible decision space, $\textbf{x}=[x_1,x_2,...,x_D]^T$ is a decision vector, and \emph{D} is the dimension size, representing the number of the decision variables involved in the problem; $f_1^{\textrm{eff}}(\textbf{x}), f_2^{\textrm{eff}}(\textbf{x}),..., f_M^{\textrm{eff}}(\textbf{x})$ are $M$ mean effective objective functions for optimization. An illustration of robust solutions is given in Fig. S.1 in the supplementary file.

For the robust order scheduling problem in the fashion industry, we set $M=2$. And $f_i(i=1, 2)$ is provided in Eqs. (\ref{eqobj1}) and (\ref{eqobj2}), hence the objectives of the problem in this paper are transformed into: $f_1^{\textrm{eff}}$ and $f_2^{\textrm{eff}}$. For the calculation of $f_i^{\textrm{eff}}$, a practical way is to generate a finite set of $H$ solutions in a randomly or structured manner, which are selected around a $\delta$-neighborhood $\mathcal{B}_\delta(\textbf{x})$ of a solution $\textbf{x}$ in the decision space; then the value of the mean effective objective function $f_i^{\textrm{eff}}$ can be calculated by averaging the function values of the $H$ neighboring solutions.

\subsection{NSJADE-Based Optimization Process}
\begin{figure}[!htbp]
\begin{minipage}[t]{1\linewidth}
\centering
\includegraphics[width=8cm]{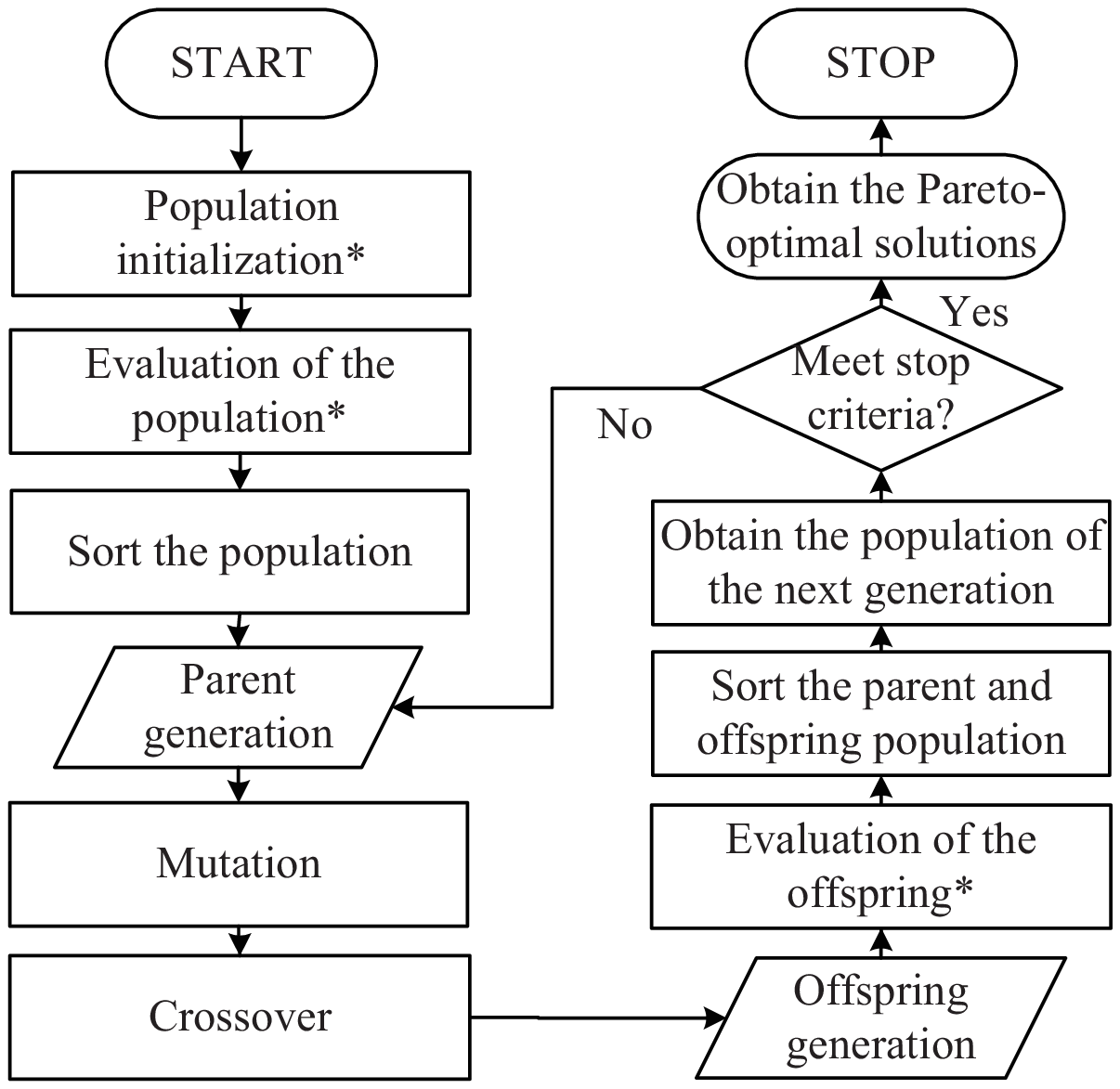}
\caption{Flowchart of the proposed NSJADE.} \label{flchart}
\end{minipage}
\end{figure}

NSJADE is developed based on two EAs: nondominated sorting genetic algorithm-II (NSGA-II) \cite{deb2002fast} and adaptive differential evolution (JADE) \cite{zhang2009jade}, which aims to combine the advantages of these two EAs. In this research, NSJADE serves as the optimization tool in the optimization process, and the flowchart of NSJADE is provided in Fig. \ref{flchart}. In the optimization process, there are three important issues we need to elaborate: encoding scheme, population initialization, and evaluation of the population. \emph{Encoding scheme} and \emph{population initialization} are related to the first box (marked by $\ast$) of the flowchart in Fig. \ref{flchart}, \emph{evaluation of the population} is related to the second and eighth boxes (marked by $\ast$) of the flowchart in Fig. \ref{flchart}.

\subsubsection{Encoding Scheme}
\begin{figure}[!htbp]
\begin{minipage}[t]{1\linewidth}
\centering
\includegraphics[width=8.8cm]{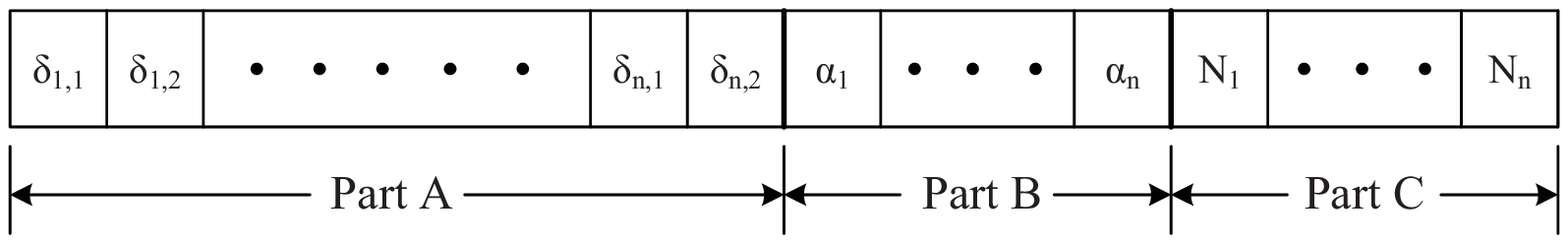}
\caption{Encoding of the individual.} \label{chro}
\end{minipage}
\end{figure}

The first step of the optimization process is to encode potential order scheduling solutions into individuals. In this research, a feasible solution needs to be able to determine the assignment of each production order to a proper production line. In addition, an individual should reflect the split information of each order, and the sequence of the orders on the same production line. Hence, each individual consists of three parts: the assignment of each production order to the production line, the split information of each order, and the sequence of the orders on the same production line. Since this research focuses on the impact of pre-production events and the uncertainties in the daily production quantity on the fashion order scheduling, to simplify the problem, we assume that during the production process, each order can be divided into at most two sub-orders. Therefore, the length of an individual is four times the number of the orders: $D=4n$. Fig. \ref{chro} illustrates the encoding of the individual. In Part A, every two bits represent the production lines to which the sub-orders of an order are assigned; and the length of Part A is $2n$. In Part B, each single bit denotes the split percentage of every order; the length of Part B is $n$. Part C assigns the label to each order, which determines the sequence of the orders on the same production line; the length of Part C is also $n$.

\begin{remark}
We assume that each order can be divided into at most 2 sub-orders, hence the dimension size of the individual $D$ equals $4n$. If we assume that an order can be split into $q$ sub-orders ($q>2$), then $D=qn+(q-1)n+n=2qn$. As the value of $q$ increases, which leads to the increase of $D$, more individuals and generations are needed for NSJADE to search for the optimal schedules. When $D$ increases to vastly more than 100 \cite{cheng2015competitive}, in order to have efficient performance, evolutionary algorithms specially designed for high-dimensional optimization problems are required.
On the other hand, it is worth pointing out that learning curve is one of the factors that affect the efficiency of apparel manufacturing. If an order is split into fewer sub-orders, the quantity of each sub-order will increase, which means the operator will repeat the production process of this product for more times, and hence improves the production efficiency.
In addition, the increase in the number of sub-orders will raise the likelihood that the neighbouring orders on the same production lines are of different product types, which then lowers the production efficiency and increases the production time of the orders.
Therefore, it is not often to split an order into many sub-orders in apparel manufacturing.
\end{remark}

\subsubsection{Population Initialization}
For the initialization of Part A, uniform random integers in the range $[1, m]$ are assigned to each dimension of Part A, where $m$ is the number of the production line.
It is worth noting that if certain lines cannot absolutely accept one type of product (the efficiency is 0 for this type of product), then related modifications should be made when initializing Part A. For example, if order 1 cannot be processed on line $P_{\tilde{m}}$ ($1\leq\tilde{m}\leq m$), then the first two bits of the encoding will be initialized in the range $[1, \tilde{m})\bigcup(\tilde{m}, m]$.
Part B shows the split percentage of each order. To simplify the optimization process, each bit of Part B is selected from $[0.2, 0.4, 0.6, 0.8]$ in a uniformly random way. For initializing Part C, each bit is assigned with a uniform random integer in the range $[1, n]$, where $n$ is the number of the production order.

\subsubsection{Evaluation of the Population}
After the population initialization, the fitness value $f_i^{\textrm{eff}}(i=1, 2)$ of each individual needs to be evaluated. To calculate $f_i^{\textrm{eff}}$, $H$ neighbouring points will be selected around the individual within a predefined range. As illustrated in Section \ref{sec1}, $\beta$ is set as the uncertainty factor of daily production quantity $\emph{Q}_\emph{jd}$. The detailed evaluation process of each individual is explained as follows, which involves six steps:

\begin{enumerate}[Step 1.]
\item Split the orders according to the split percentage in Part B of the individual's encoding if the orders are scheduled to be processed on different production lines in terms of Part A.
\item According to Part A of the individual's encoding, assign the orders or sub-orders to the production lines.
\item Based on Part C of the individual's encoding, determine the sequence of the orders assigned to the same production line.
\item Generate $H$ neighbouring points of the individual, and for each neighbouring point, the daily production quantity $\emph{Q}_\emph{jd}$ can be calculated according to Line 11 in Algorithm 1.
\item Calculate the fitness value $f_i(i=1, 2)$ of each neighbouring point with Eqs. (\ref{eqobj1}) and (\ref{eqobj2}).
\item Average the fitness values of $H$ neighbouring points, and the fitness value $f_i^{\textrm{eff}}(i=1, 2)$ of the individual can be obtained.
\end{enumerate}

After initializing the population, several operations, e.g., fast nondominated sorting, crowding-distance assignment, mutation, crossover, and selection, are carried out. Because of the page limit, the details of these operations are provided in Section S.I in the supplementary file.

\section{Experiments and Analysis of Results}

\subsection{Experimental Information}
Because of the page length limit, the experimental setup, including the information of the test data, the production orders and the production lines, is given in Section S.II in the supplementary file.

The parameters of NSJADE are set as the same as that in JADE \cite{zhang2009jade}. It should be mentioned that in this paper, we do not use the maximum number of function evaluations (MAX\_FES) to control the termination of NSJADE. The reason is that MAX\_FES relies on the value of $H$, which is the number of the neighbouring points of an individual. In order to simply control the termination of NSJADE, we utilize the maximum number of generations $\emph{G}_{\emph{max}}$ as the stopping criterion; and $\emph{G}_{\emph{max}}=\emph{D}\cdot\xi$, where \emph{D} is the dimension size, and $\xi$ is a predefined parameter, which controls the evolution generations of the algorithm. The value of $\xi$ is set as 10 according to the analysis in Section S.III in the supplementary file. It is noticed that a differential evolution (DE) algorithm is employed as the search engine of NSJADE; and the population size of DE is recommended as $5\cdot\emph{D}$ \cite{das2011differential}. Therefore, the population size in this paper is $\emph{NP}=400$, since the dimension size of the problem investigated is 80. In the experiments, the number of the neighbouring points is set as $H=5$, and the uncertainty factor of daily production quantity is set as $\beta=0.2$. NSJADE is conducted for 30 runs for eliminating discrepancy, and the PFs are sorted out from the solutions obtained after 30 runs.

\subsection{Significance of Considering Pre-Production Events}\label{secc}
One of the contributions of this research is to consider pre-production events when making the order schedules in the fashion industry. In this subsection, two groups of experiments will be conducted to show the significance of including pre-production events.
\begin{figure}
\centering
\subfigure[]{
\begin{minipage}[b]{0.22\textwidth}
\includegraphics[width=1\textwidth]{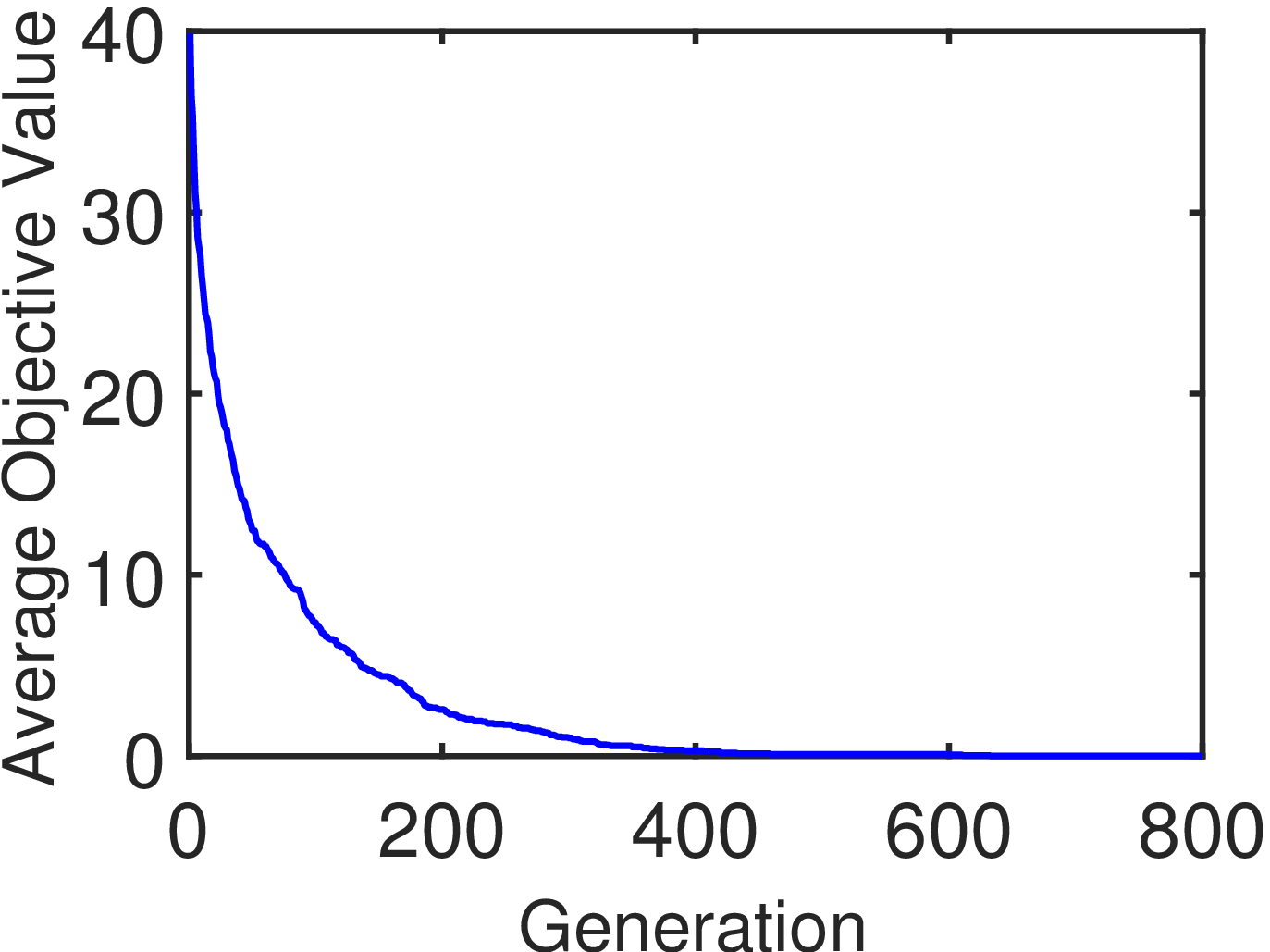}
\end{minipage}\label{soplot}
}
\subfigure[]{
\begin{minipage}[b]{0.22\textwidth}
\includegraphics[width=1\textwidth]{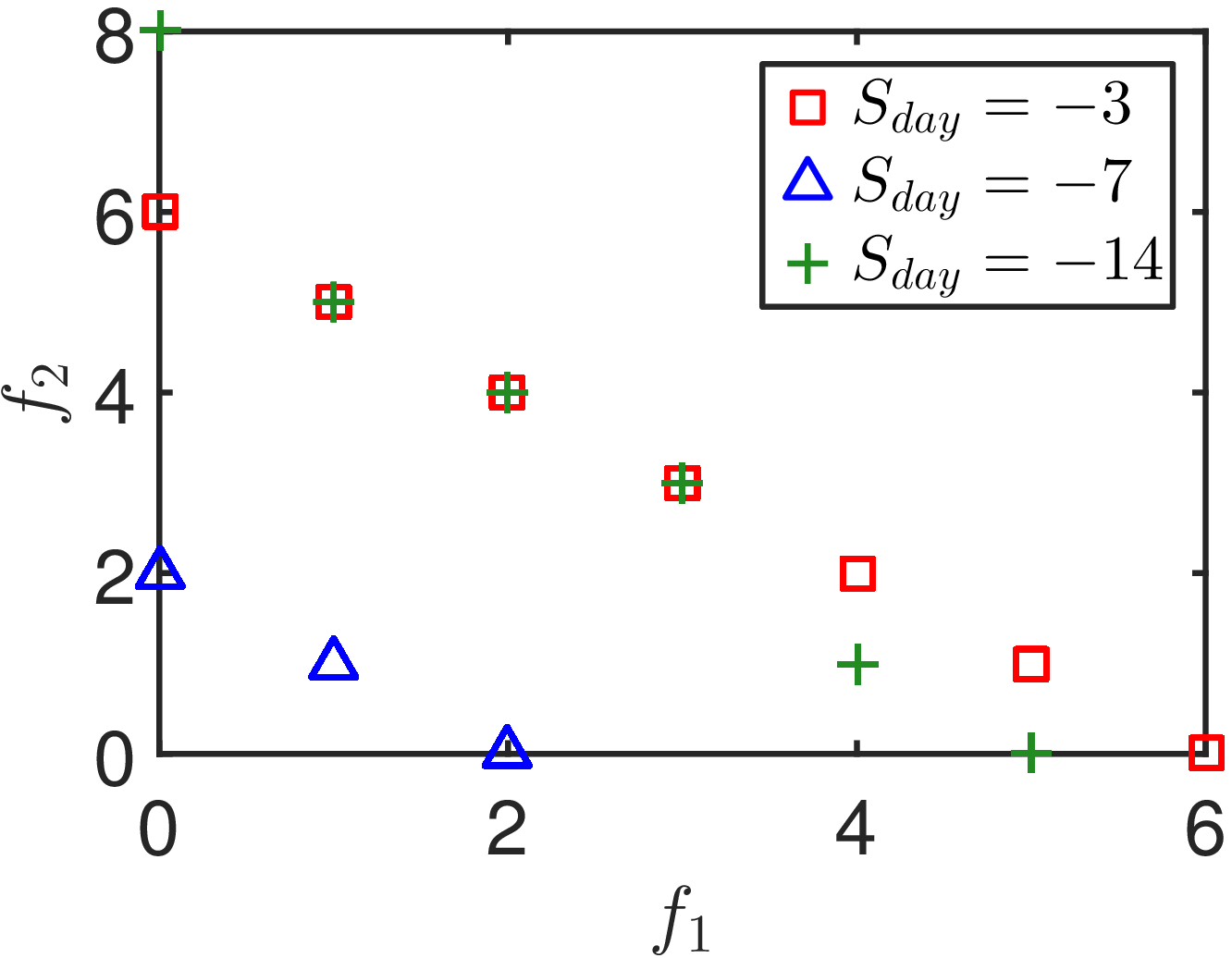}
\end{minipage}\label{3pf}
}
\caption{The illustration of the significance of considering pre-production events. (a) The evolution of the average objective value of 30 runs for the order scheduling problem optimized by JADE when pre-production events are not considered; (b) The PFs obtained by NSJADE when $\emph{S}_\emph{day}=-3$, $\emph{S}_\emph{day}=-7$, and $\emph{S}_\emph{day}=-14$, respectively.} \label{pretwo}
\end{figure}

In the first group of experiments, no pre-production events are taken into consideration, which means that all the orders are ready for production after the production begins. The problem is transformed into a single-objective optimization problem, and the only objective is to minimize the total tardiness of all orders. A single-objective evolutionary algorithm JADE \cite{zhang2009jade} is utilized to search the optimal solution for order schedules. We provide the evolution of the average objective value of 30 runs for the problem optimized by JADE when pre-production events are not considered in Fig. \ref{soplot}. From Fig. \ref{soplot}, it shows that the average objective value drops to 0 within 800 generations, which means that the schedule with no tardiness can be achieved easily.

In the second group of experiments, pre-production events are taken into consideration when making the schedule. In real-world production, based on the progress of the orders' pre-production events, planners make or update order schedules at regular intervals before the production begins. Here, we make the schedules 3 days, 7 days, and 14 days before the production begins, i.e., $\emph{S}_\emph{day}=-3$, $\emph{S}_\emph{day}=-7$, and $\emph{S}_\emph{day}=-14$. The problem is optimized by our proposed NSJADE and the PFs are provided in Fig. \ref{3pf}. The results illustrate that the schedule with no pre-production event clashes and no tardiness cannot be obtained when pre-production events are included in the order scheduling. Therefore, it can be concluded that pre-production events have a big impact on the order scheduling problems in the fashion industry, which should not be neglected in the research of order scheduling.

\begin{remark}
In multi-objective optimization, the ideal solution denotes an array of the lower bound of all objective functions (for minimization problems) \cite{deb2001multi}. For the order scheduling problem in this paper, the ideal solution is $[0, 0]$, which means the total tardiness of all orders is 0, and the total pre-production event clashes of all orders are also 0. However, in real-world production, multiple parties, e.g., suppliers, manufacturers and customers, need to collaborate with each other to complete a pre-production event. And negotiations among them are time-consuming and uncertain. In addition, some events are closely linked, which means the only requirement to begin an event is that another event must be finished.
Therefore, when making the schedules, the total pre-production event clashes of all orders mostly exist. If we hope that the total clashes become 0, we need to postpone the starting date of certain orders when making the schedule. However, the late starting date of an order might defer its finishing date, which means the tardiness happens. So minimizing the total tardiness of all orders and minimizing the total pre-production event clashes of all orders are two conflicting objectives of the order scheduling problem in this paper.
By means of NSJADE, a set of schedules which balance these two objectives can be provided.
\end{remark}

\begin{remark}
It is worth mentioning that the PFs obtained by NSJADE when $\emph{S}_\emph{day}=-3$, $\emph{S}_\emph{day}=-7$, and $\emph{S}_\emph{day}=-14$ are different. This is because the progress of the orders' pre-production events will keep updating when the production is approaching, which then influences the present conservative starting date $\emph{C}_\emph{day,j}$ of each order. According to Eq. (\ref{eqobj2}), $\emph{C}_\emph{day,j}$ is involved in the second objective; therefore, the PFs obtained by NSJADE may be in different shapes when $\emph{S}_\emph{day}$ is distinct. From Fig. \ref{figH}, it can be noticed that the robust PF for $\emph{S}_\emph{day}=-7$ offers the best result when compared with those for $\emph{S}_\emph{day}=-3$ and $\emph{S}_\emph{day}=-14$; and the higher flexibility (i.e., $\emph{S}_\emph{day}=-14$) in the time schedule does not increase the chance for less clashes, which is also caused by the present conservative starting date $\emph{C}_\emph{day,j}$ of each order. By taking a look at $O_{1}$, $O_{2}$, $O_{3}$, $O_{7}$, $O_{9}$, $O_{11}$, $O_{12}$ and $O_{17}$ in Table S.1 in the supplementary file, we find that $\emph{C}_\emph{day,j}$ for $\emph{S}_\emph{day}=-14$ (i.e., 6 or 11) is larger than that for $\emph{S}_\emph{day}=-3$ (i.e., 0) and $\emph{S}_\emph{day}=-7$ (i.e., 0). This means these orders can only be arranged on the 6th or 11th day after the production begins when $\emph{S}_\emph{day}=-14$, and we can say more restrictions are put on scheduling these orders, which also explains why the higher flexibility in the time schedule does not decrease the clashes.
\end{remark}

\subsection{Comparison of Non-Robust and Robust Order Schedules by NSJADE}\label{secd}
\begin{figure}
\centering
\subfigure[$\emph{S}_\emph{day}=-3$]{
\begin{minipage}[b]{0.19\textwidth}
\includegraphics[width=0.95\textwidth]{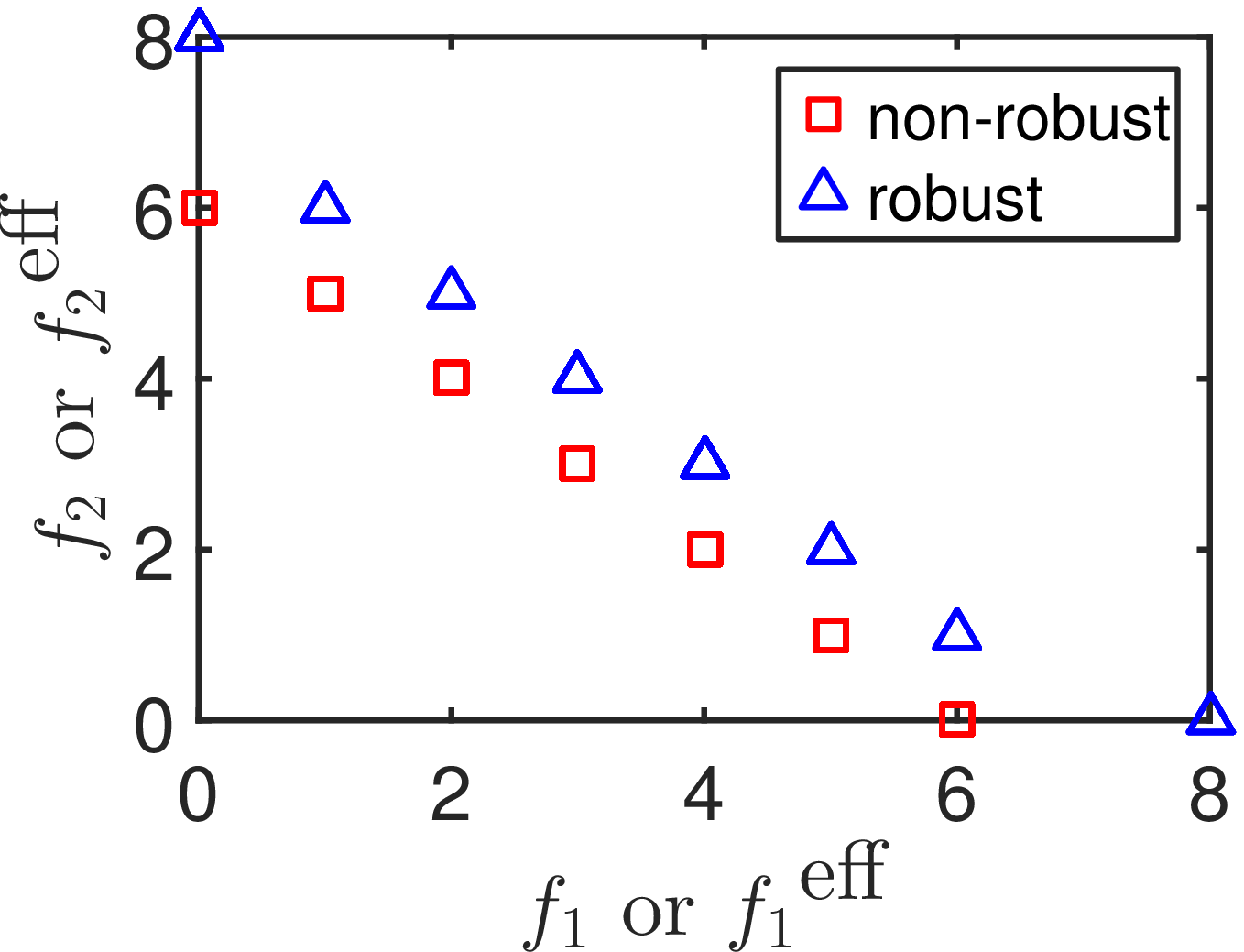}
\end{minipage}
}
\subfigure[$\emph{S}_\emph{day}=-7$]{
\begin{minipage}[b]{0.19\textwidth}
\includegraphics[width=0.95\textwidth]{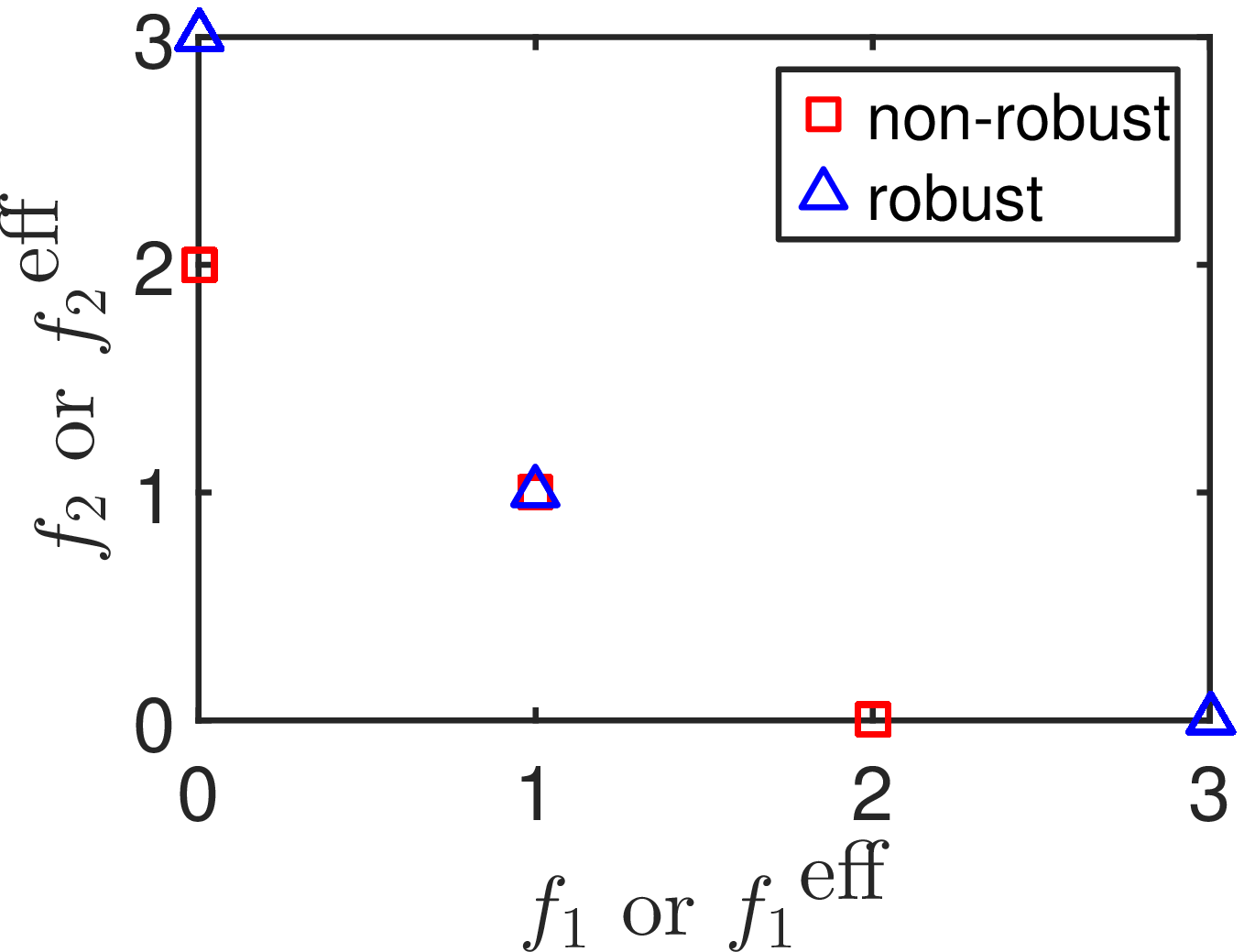}
\end{minipage}
}
\subfigure[$\emph{S}_\emph{day}=-14$]{
\begin{minipage}[b]{0.19\textwidth}
\includegraphics[width=0.95\textwidth]{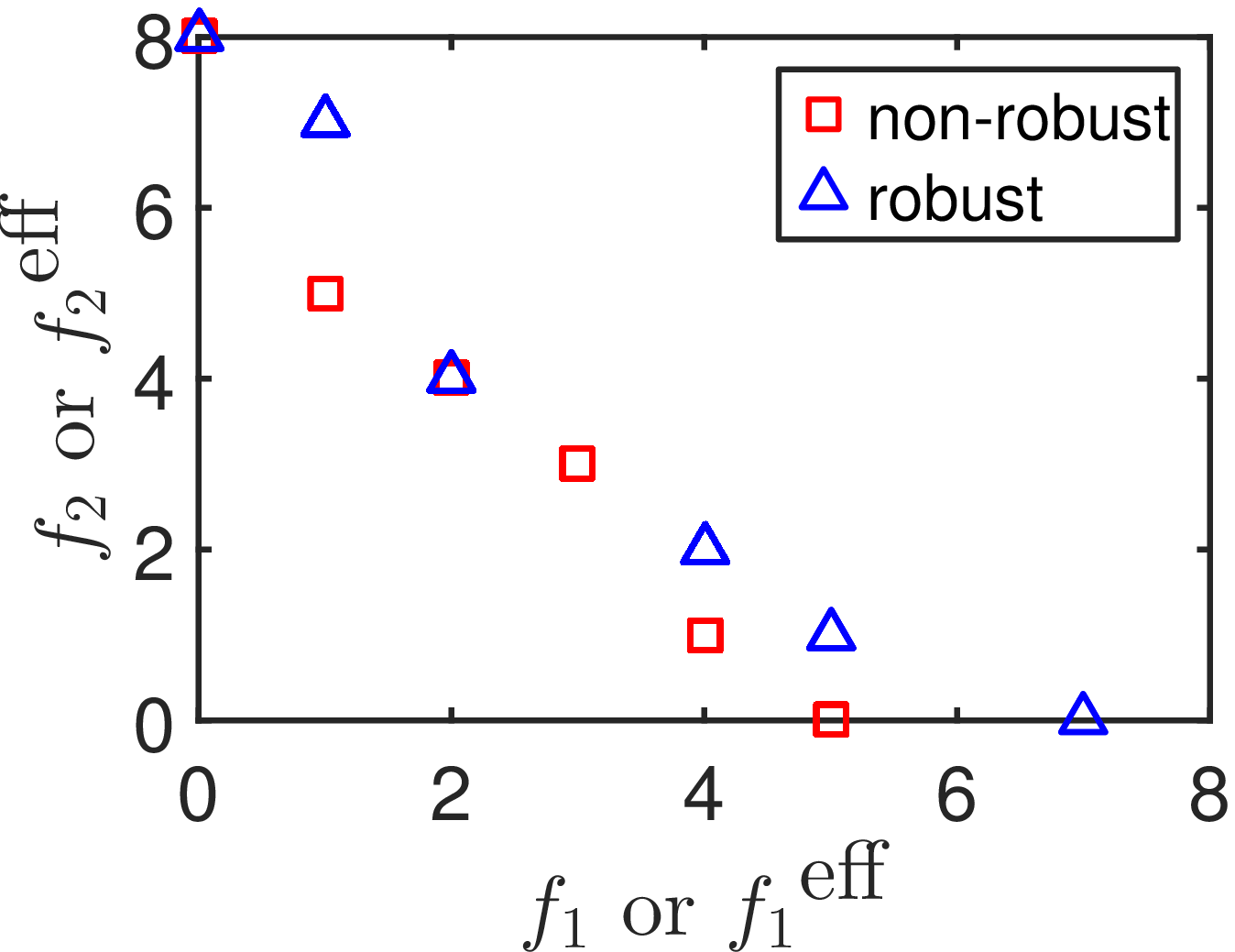}
\end{minipage}\label{expl}
}
\caption{Comparisons of the PFs of non-robust and robust order schedules obtained by NSJADE. (a) $\emph{S}_\emph{day}=-3$; (b) $\emph{S}_\emph{day}=-7$; (c) $\emph{S}_\emph{day}=-14$.} \label{figH}
\end{figure}

In this subsection, we will compare the non-robust and robust order schedules obtained in the experiments, which aims at illustrating the significance of introducing robust multi-objective optimization into order scheduling problems in the fashion industry. The results are provided in Fig. \ref{figH}.

From Fig. \ref{figH}, it can be observed that the schedules obtained achieve a balance between the two objectives, i.e., minimizing the total tardiness of all orders and minimizing the total pre-production event clashes of all orders.
%The results unveil that orders' pre-production events play an important role in the order scheduling problems in the fashion industry, and cannot be neglected. In addition,
The results unveil that when the daily production quantity is not fixed, which means the uncertainty is considered, the PFs shown in Fig. \ref{figH} are different from that of the cases without uncertainty. The differences appear due to the uncertain daily production quantity during the production.
In robust schedules, the daily production quantity of each order is allowed to vary within a certain range, and we can say that the robust schedules (represented by blue triangles in Fig. \ref{figH}) are less sensitive to the variation of each order's daily production quantity. As a result, robust order schedules can be shifted less often after the production starts than non-robust ones, which saves labor cost and enhances the production efficiency. In addition, based on the robust order schedules obtained by NSJADE, planners can pay close attention to the unfinished pre-production events as early as possible.

In the following, we randomly take two of the Pareto solutions obtained in a non-robust scenario as an example, and investigate how the solutions will be affected when uncertainty is considered. In detail, a solution A ($(f_1, f_2)=(1, 5)$ in the objective space) and a solution B ($(f_1, f_2)=(5, 0)$ in the objective space) are selected from the PF by NSJADE without uncertainty in Fig. \ref{expl}. The schedules represented by solution A and solution B are marked as schedule SA and schedule SB. To save space, we only list the details of schedule SA in Table \ref{expr2}; the details of schedule SB are provided in Table S.3 in the supplementary file. The figures in the parentheses denote the specific order size after order split.
\begin{table}[!htbp]
\scriptsize
\centering
\caption{The details of the order assignments on 6 production lines in schedule SA.}\label{expr2}
\begin{tabular}{|c|c|}
\hline
\tabincell{c}{\textbf{Production} \\ \textbf{Line No.}} & \textbf{Order Assignments} \\\hline
 1 & \tabincell{c}{$O_6(400)$, $O_5(600)$, $O_{10}(312)$, $O_2(420)$, $O_1(348)$,\\ $O_9(800)$, $O_{20}(2400)$, $O_{13}(240)$}   \\\hline
 2 & \tabincell{c}{$O_4(500)$, $O_{10}(468)$, $O_{18}(320)$, $O_2(280)$, $O_1(522)$,\\ $O_3(320)$, $O_{17}(800)$, $O_{20}(600)$}  \\\hline
 3 & \tabincell{c}{$O_5(400)$, $O_{15}(600)$, $O_7(480)$, $O_3(480)$, $O_{14}(2000)$,\\ $O_{11}(200)$, $O_{13}(160)$}  \\\hline
 4 & $O_6(600)$, $O_{15}(400)$, $O_7(320)$, $O_{18}(480)$  \\\hline
 5 & $O_{16}(300)$, $O_{12}(400)$, $O_8(510)$, $O_{11}(800)$, $O_{19}(420)$  \\\hline
 6 & $O_{16}(200)$, $O_{12}(600)$, $O_8(340)$, $O_{19}(280)$  \\\hline
\end{tabular}
\end{table}

In schedule SA, three orders have the pre-production event clashes or delay in delivery: $O_7$ starts 1 day earlier before all the pre-production events are ready, and finishes 1 day later than its due date; $O_{12}$ and $O_{16}$ start 3 days and 1 day earlier before all the pre-production events are ready, respectively.
When uncertainty is introduced into schedule SA, we calculate the updated objective values $[f_1^{\textrm{eff}}, f_2^{\textrm{eff}}]$, and get $[f_1^{\textrm{eff}}, f_2^{\textrm{eff}}]=[4, 5]$. Three more orders (i.e., $O_2$, $O_{17}$ and $O_{19}$) encounter the delay in delivery besides $O_7$, which leads to the increase of the first objective.
Similarly, when uncertainty is introduced into schedule SB, we get $[f_1^{\textrm{eff}}, f_2^{\textrm{eff}}]=[6, 0]$. One more order (i.e., $O_{7}$) encounters the delay in delivery besides $O_9$ and $O_{12}$, which leads to the increase of the first objective.
It can be observed that uncertainty has a big impact on the schedules, and it is meaningful to consider uncertainty in the order scheduling problem.

\subsection{Effect of $\beta$ on Robust Order Scheduling}
\begin{figure}
\centering
\subfigure[$\emph{S}_\emph{day}=-3$]{
\begin{minipage}[b]{0.19\textwidth}
\includegraphics[width=0.95\textwidth]{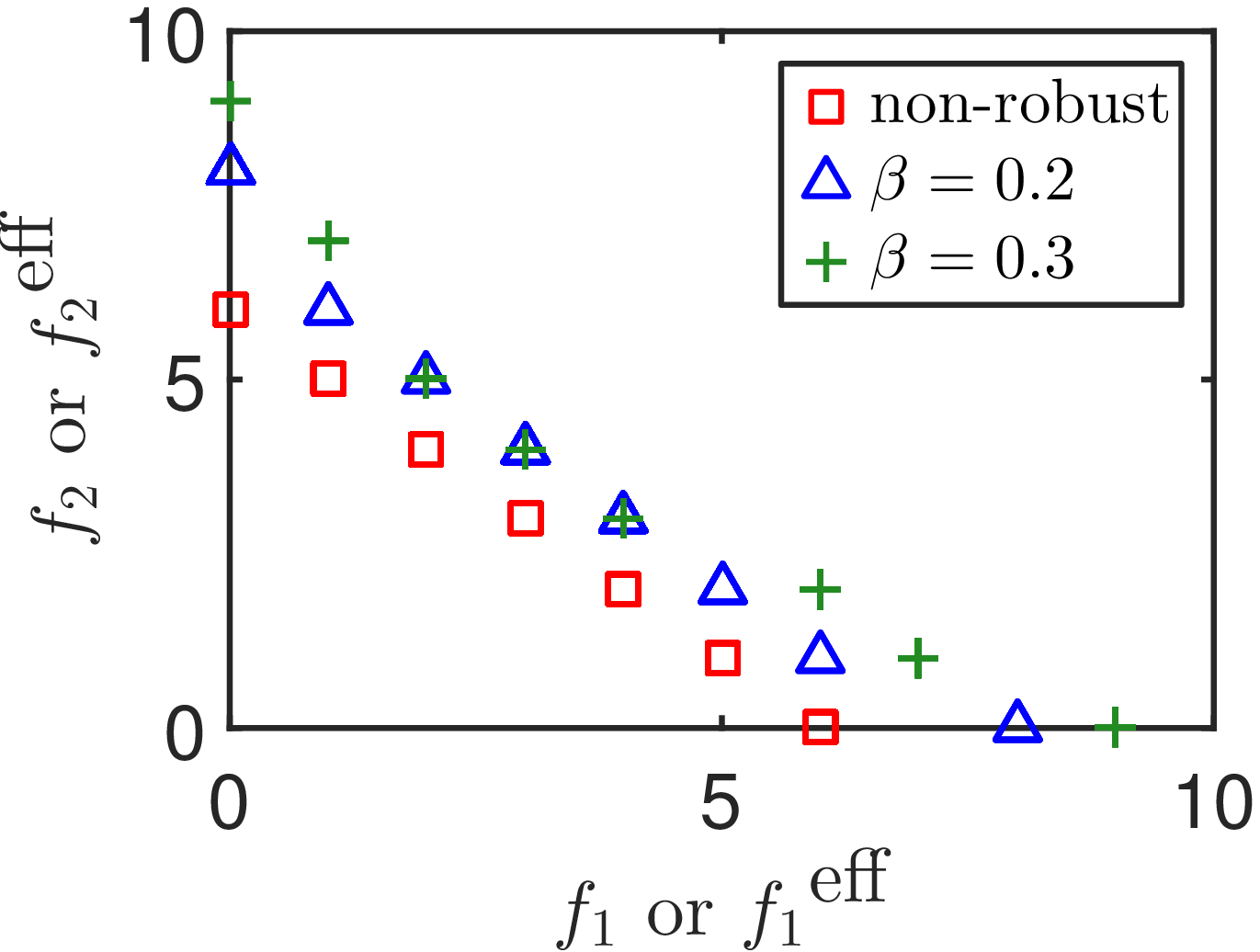}
\end{minipage}
}
\subfigure[$\emph{S}_\emph{day}=-7$]{
\begin{minipage}[b]{0.19\textwidth}
\includegraphics[width=0.95\textwidth]{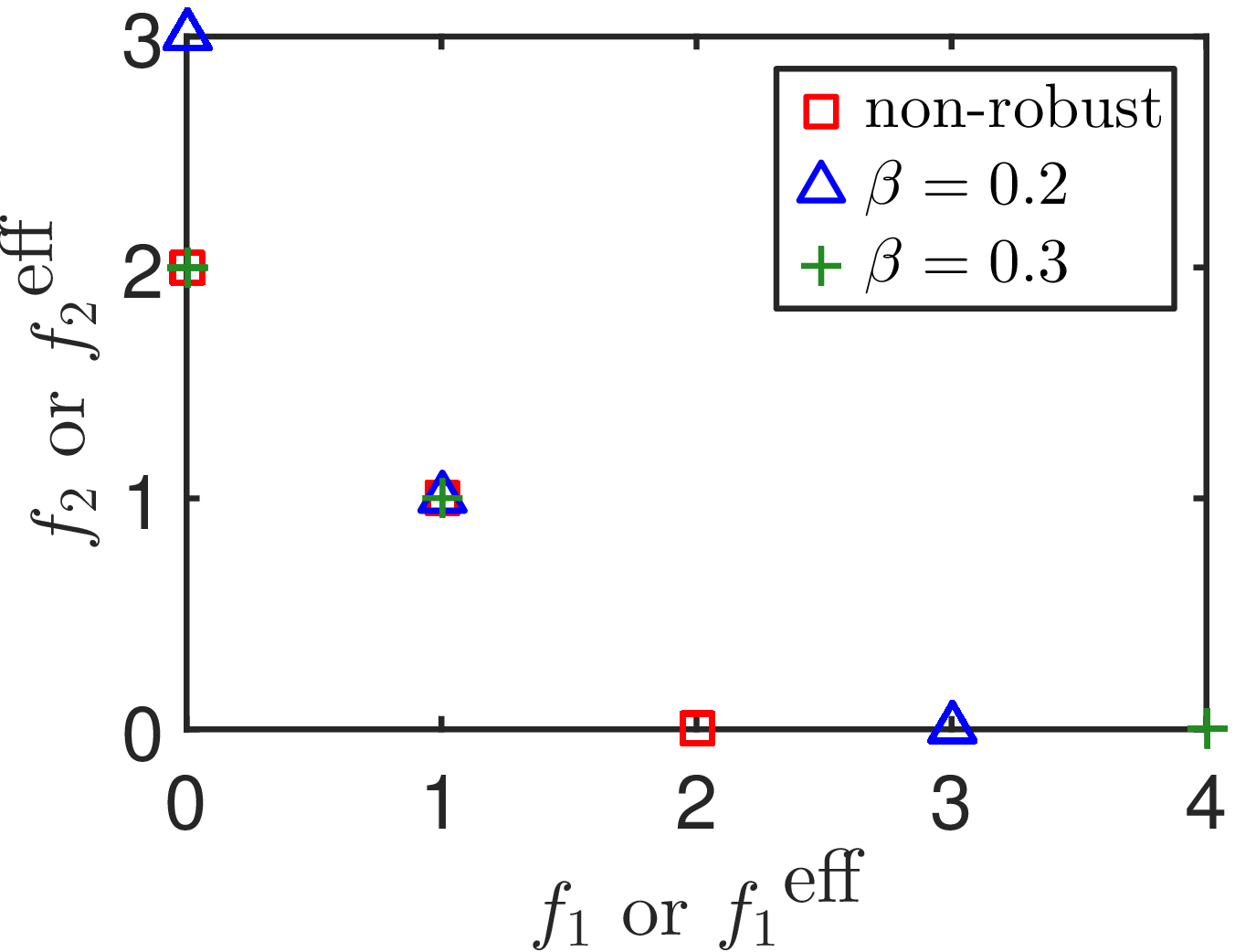}
\end{minipage}
}
\subfigure[$\emph{S}_\emph{day}=-14$]{
\begin{minipage}[b]{0.19\textwidth}
\includegraphics[width=0.95\textwidth]{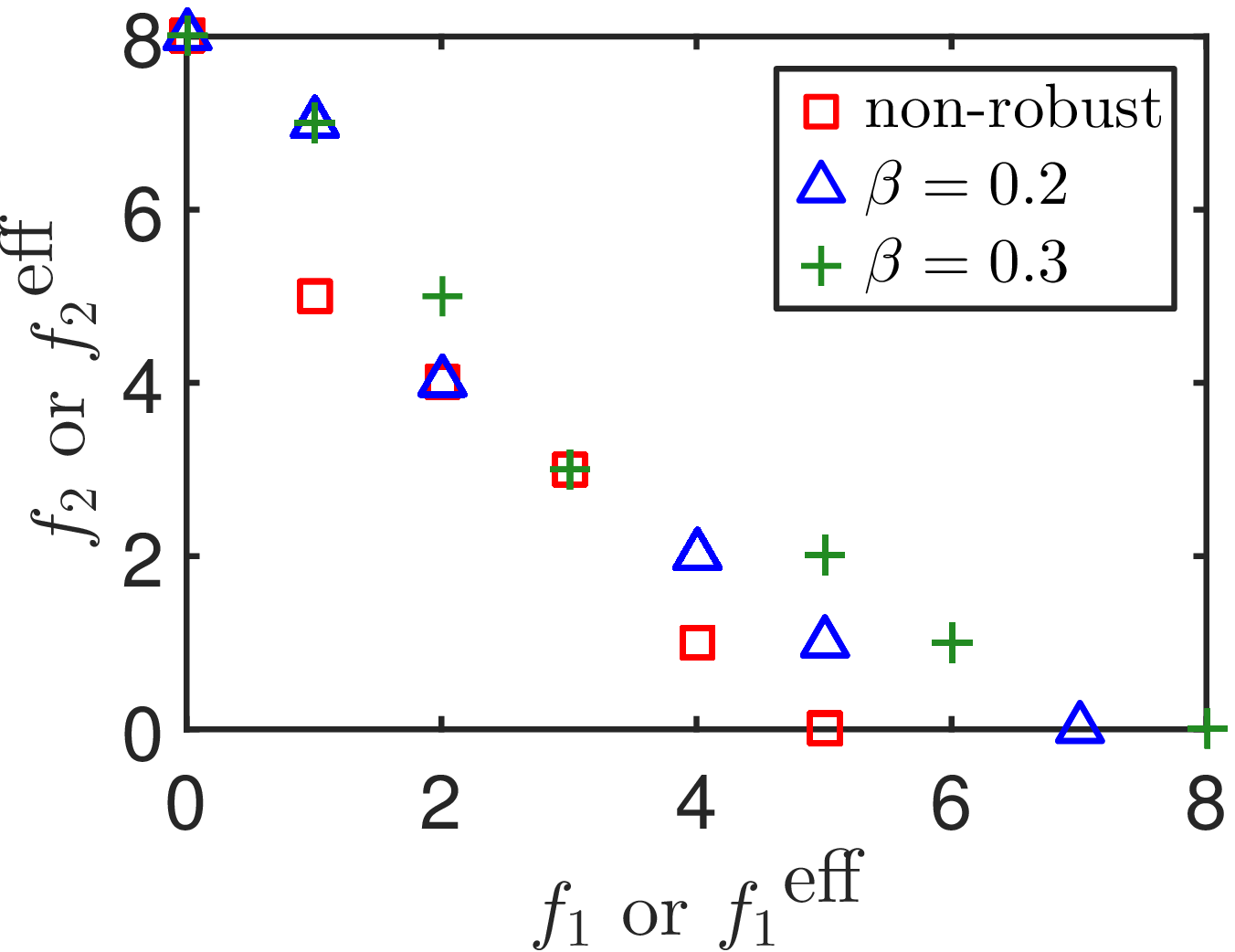}
\end{minipage}
}
\caption{Comparisons of the PFs obtained by NSJADE with different $\beta$. (a) $\emph{S}_\emph{day}=-3$; (b) $\emph{S}_\emph{day}=-7$; (c) $\emph{S}_\emph{day}=-14$.} \label{figbeta}
\end{figure}

In this subsection, we will study the impact of the uncertainty factor $\beta$ on robust order scheduling. $\beta$ is set as $\beta=[0.2, 0.3]$, and the results are displayed in Fig. \ref{figbeta}. From Fig. \ref{figbeta}, it can be observed that as $\beta$ increases, the shift in the PF moves away from the original PF, i.e., uncertainty is not considered. This phenomenon is natural, since the increase of $\beta$ brings more uncertainties in the daily production quantity of each order, which then causes more differences from the original PF.

\subsection{Effect of $H$ on Robust Order Scheduling}
\begin{figure}
\centering
\subfigure[$\emph{S}_\emph{day}=-3$]{
\begin{minipage}[b]{0.19\textwidth}
\includegraphics[width=0.95\textwidth]{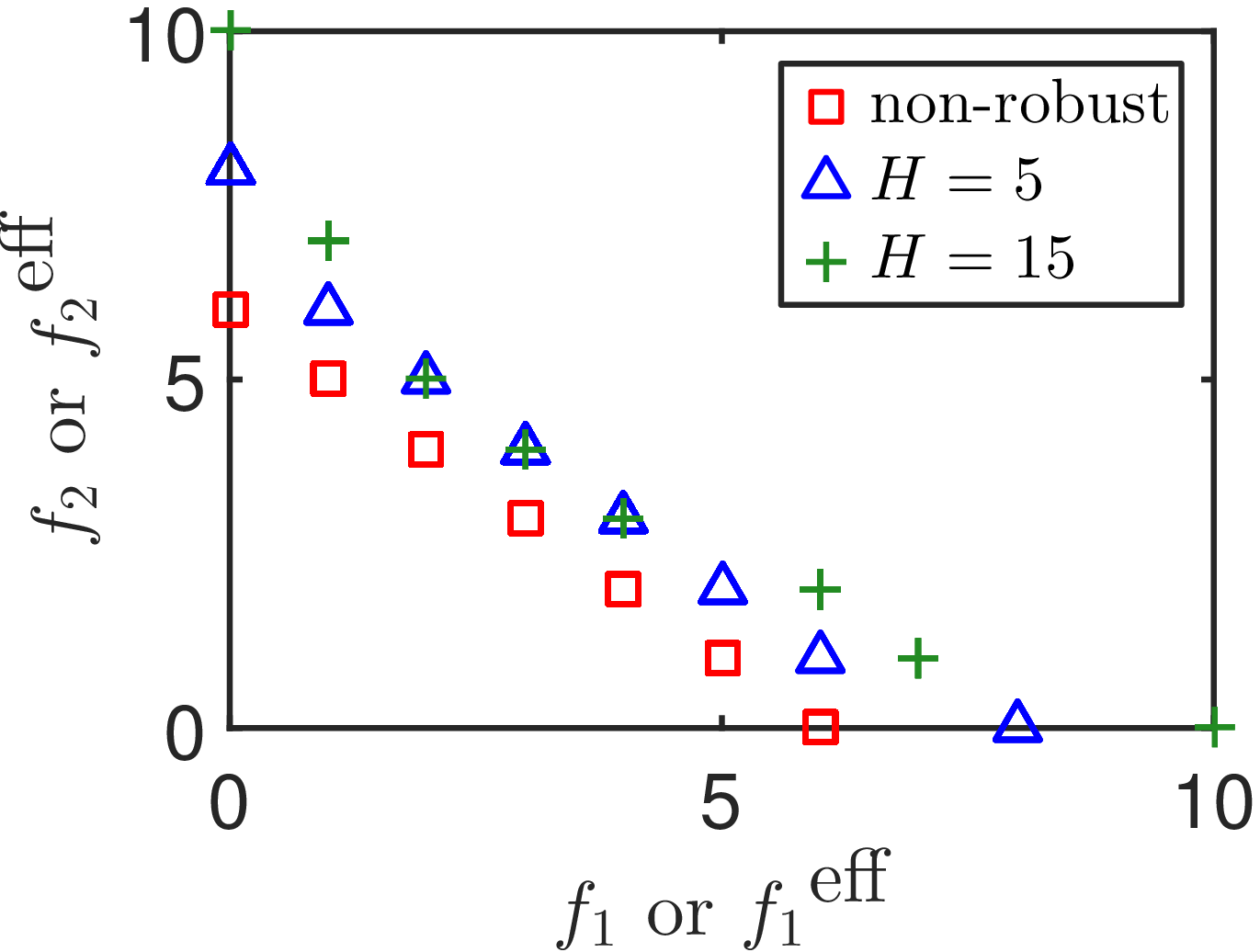}
\end{minipage}
}
\subfigure[$\emph{S}_\emph{day}=-7$]{
\begin{minipage}[b]{0.19\textwidth}
\includegraphics[width=0.95\textwidth]{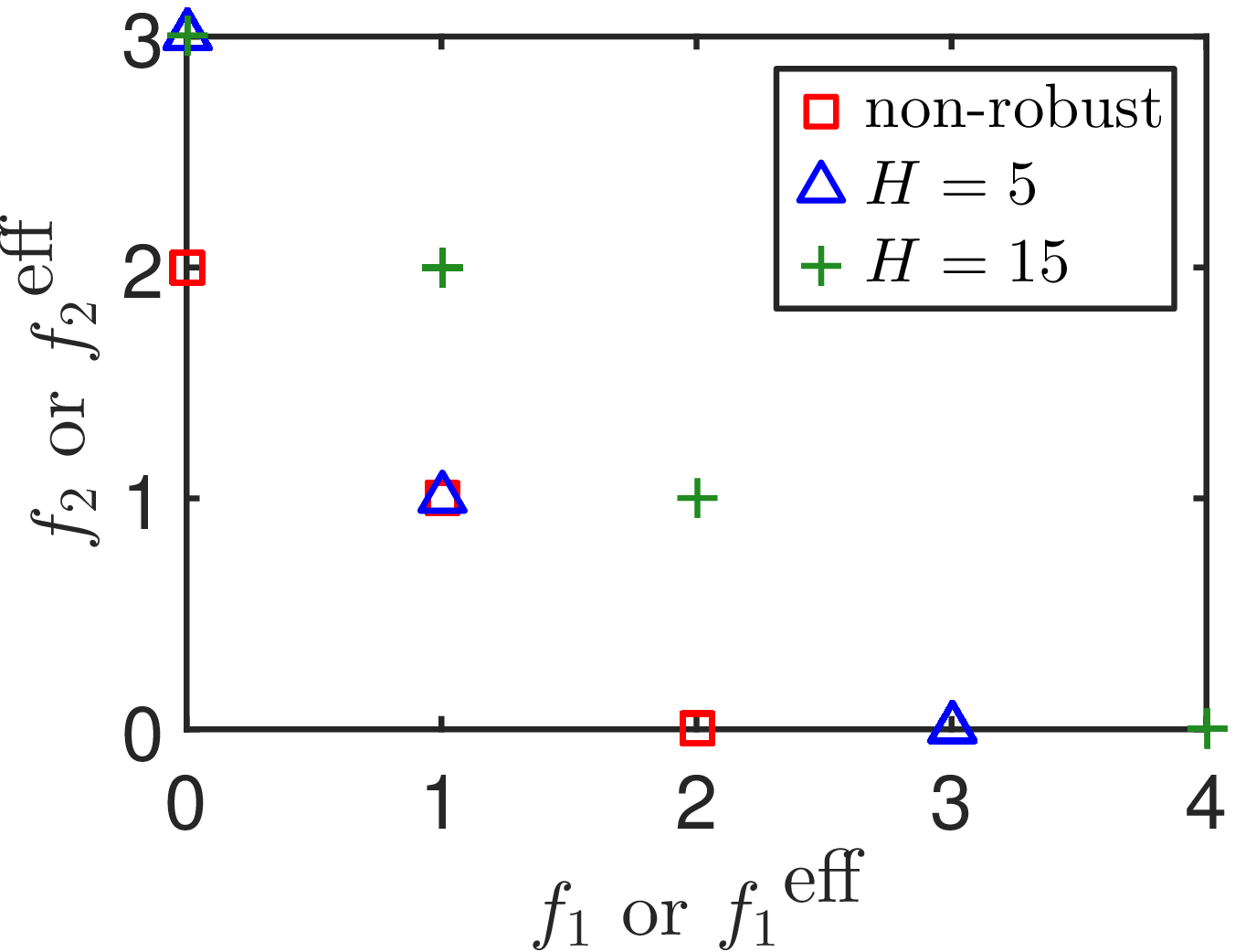}
\end{minipage}
}
\subfigure[$\emph{S}_\emph{day}=-14$]{
\begin{minipage}[b]{0.19\textwidth}
\includegraphics[width=0.95\textwidth]{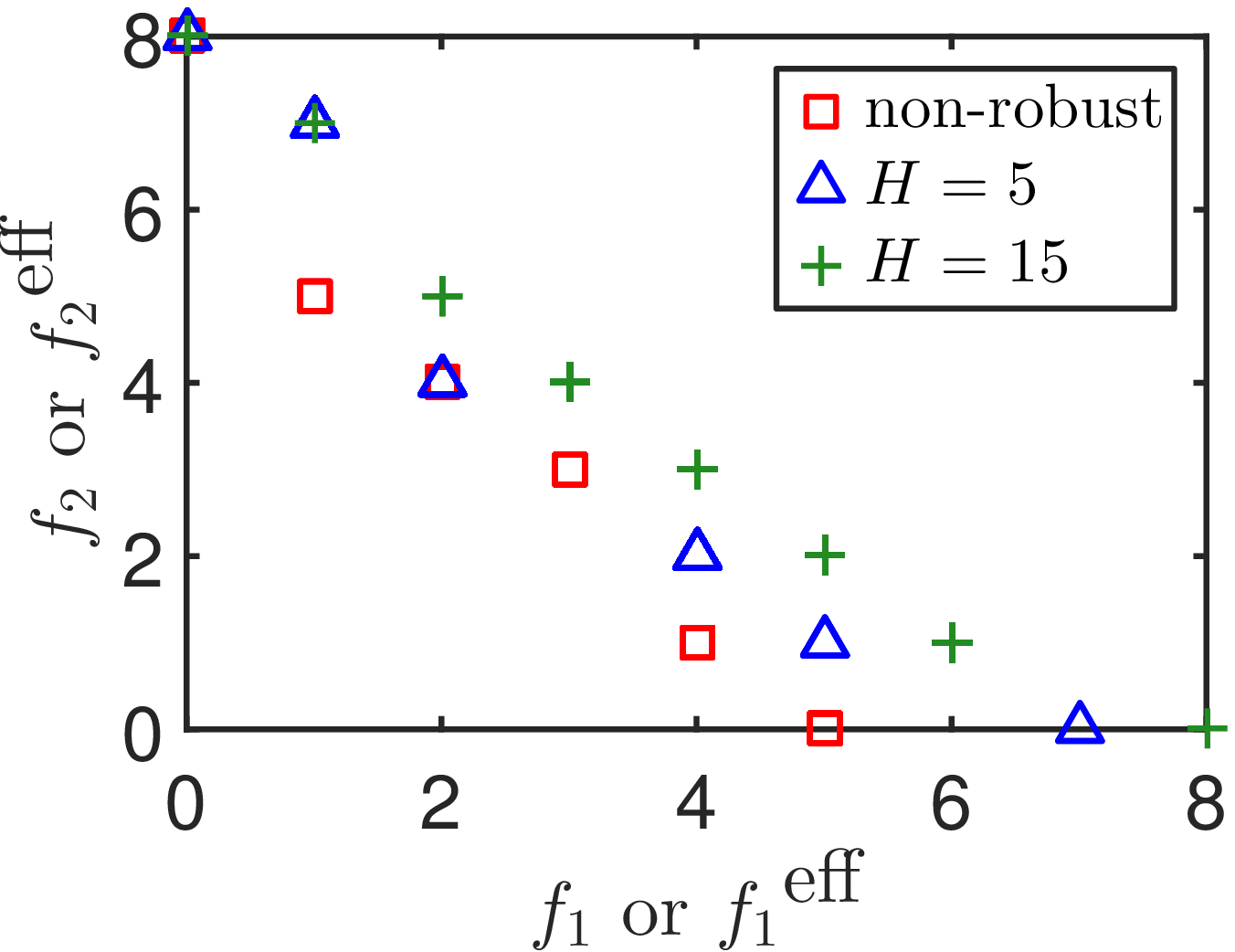}
\end{minipage}
}
\caption{Comparisons of the PFs obtained by NSJADE under different $H$. (a) $\emph{S}_\emph{day}=-3$; (b) $\emph{S}_\emph{day}=-7$; (c) $\emph{S}_\emph{day}=-14$.} \label{figH15}
\end{figure}

As described in Section \ref{defH}, $H$ neighbouring solutions are selected to compute the mean effective objective function $f_i^{\textrm{eff}}$. Intuitively, if more neighbouring solutions are chosen, the objective values will be closer to the true average values.
In the previous experiments of this section, the value of $H$ is set as 5. Here, we depict the effect of using different values of $H$ on robust order scheduling problems in Fig. \ref{figH15}. From Fig. \ref{figH15}, it can be observed that despite having a smaller computational time, the mean effective front using a small $H$ overestimates the true robust front. However, compared with $H=15$, the mean effective front with $H=5$ can also achieve a satisfactory approximation. In applications, users can choose the values of $H$ according to practical situations.

\subsection{Comparison with Existing MOEAs}
\begin{figure}
\centering
\subfigure[$\emph{S}_\emph{day}=-3$]{
\begin{minipage}[b]{0.19\textwidth}
\includegraphics[width=0.95\textwidth]{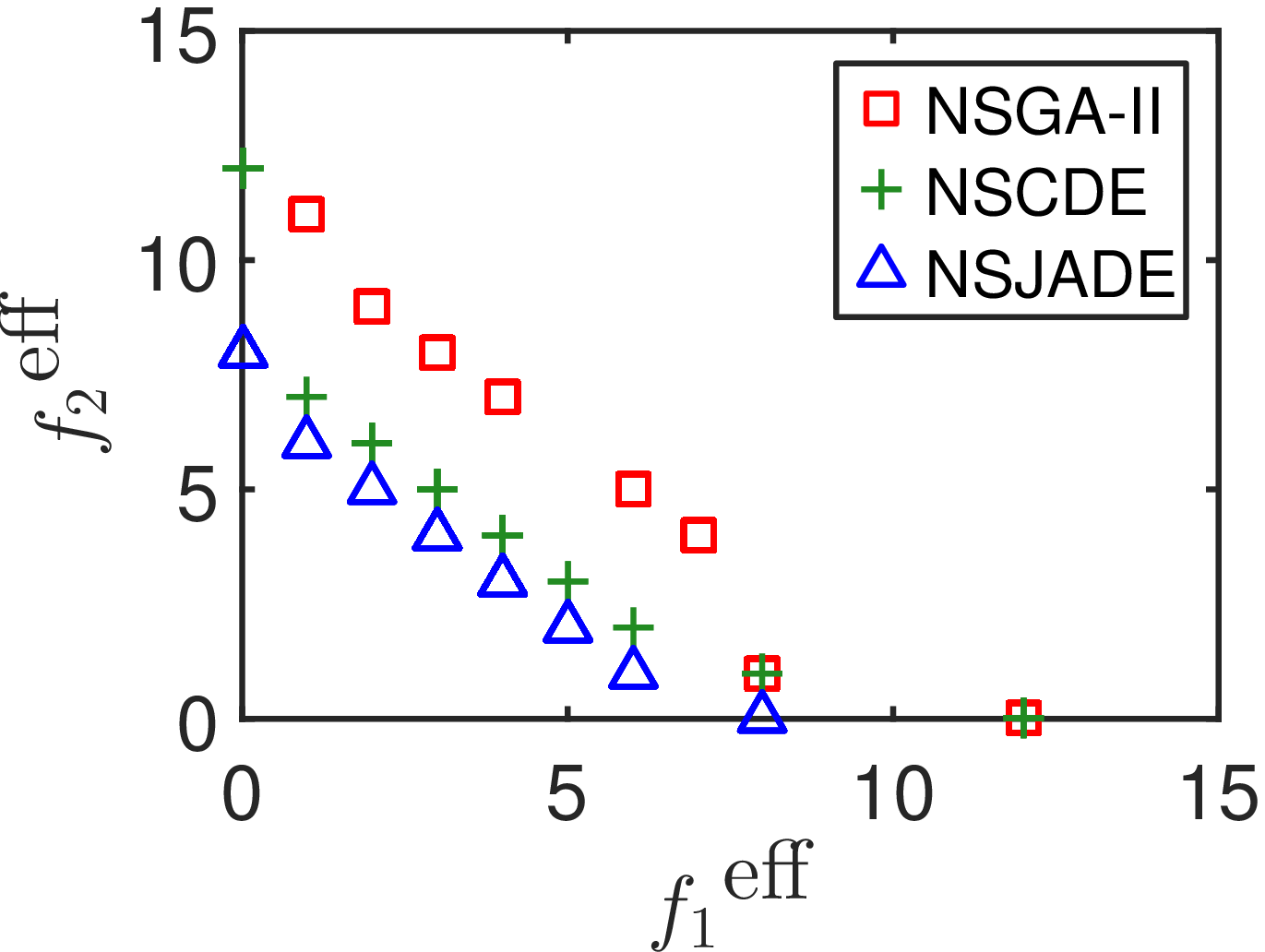}
\end{minipage}
}
\subfigure[$\emph{S}_\emph{day}=-7$]{
\begin{minipage}[b]{0.19\textwidth}
\includegraphics[width=0.95\textwidth]{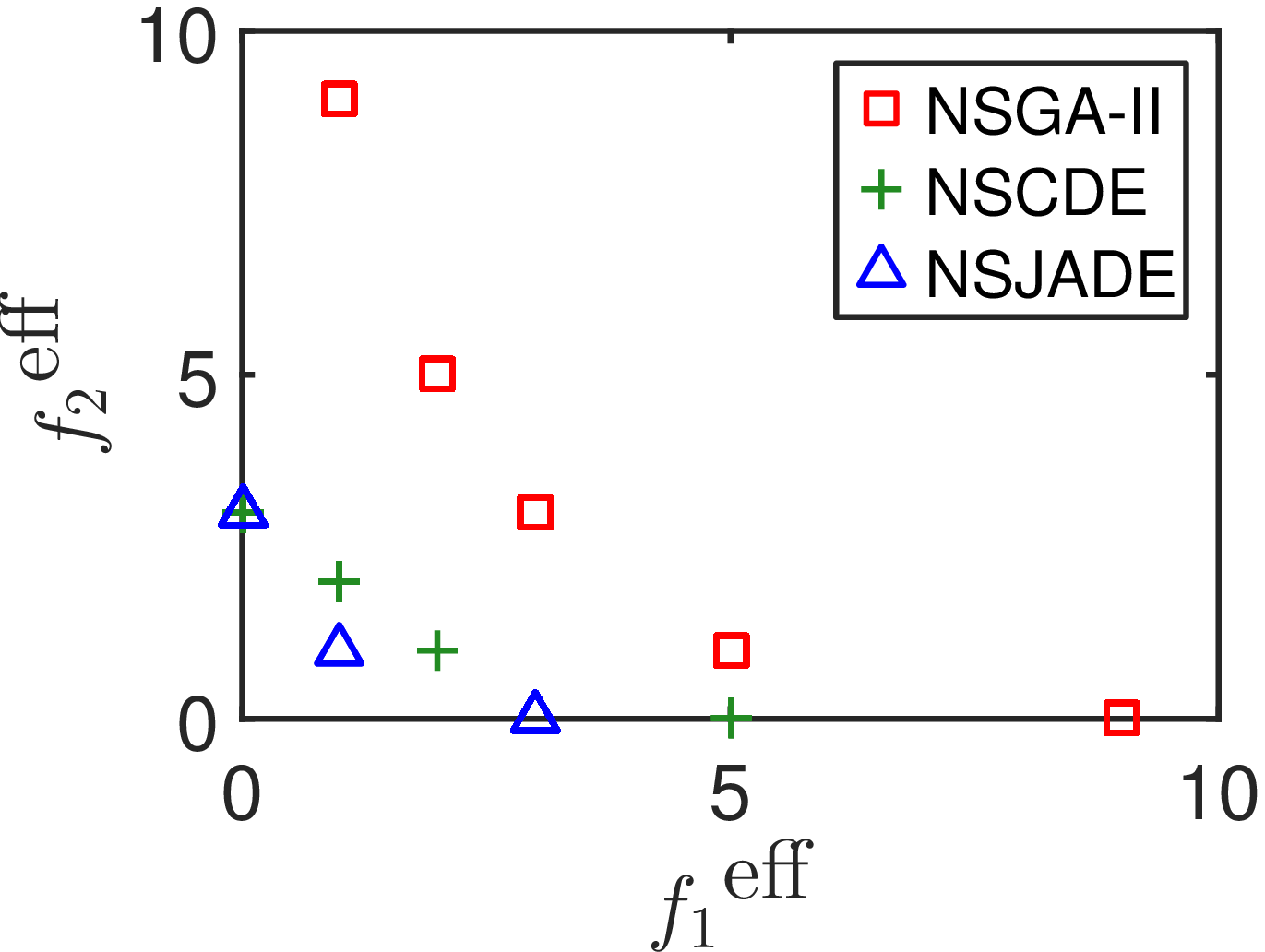}
\end{minipage}
}
\subfigure[$\emph{S}_\emph{day}=-14$]{
\begin{minipage}[b]{0.19\textwidth}
\includegraphics[width=0.95\textwidth]{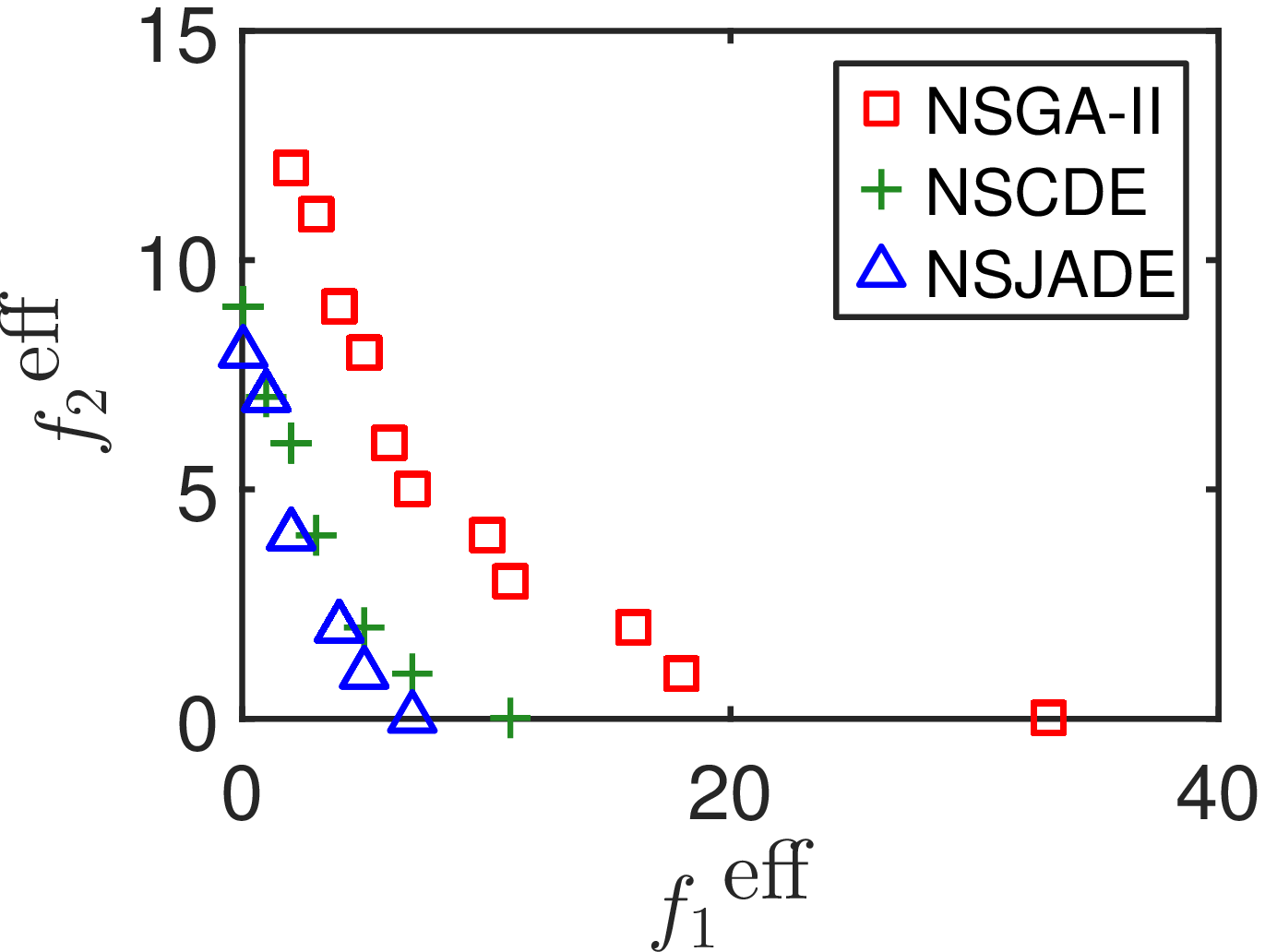}
\end{minipage}
}
\caption{Comparisons of the PFs obtained by NSGA-II, NSCDE, and NSJADE. (a) $\emph{S}_\emph{day}=-3$; (b) $\emph{S}_\emph{day}=-7$; (c) $\emph{S}_\emph{day}=-14$.} \label{figcomp}
\end{figure}

In order to show the superiority of NSJADE, we compare it with two existing MOEAs: NSGA-II \cite{deb2002fast} and NSCDE \cite{tang2013multiobjective}. The results are given in Fig. \ref{figcomp}. In this paper, the computational resource is limited (i.e., 800 generations) for each algorithm. Therefore, we need to select the algorithm that is powerful enough to obtain the solutions with better convergence performance with limited computational resource. It can be observed from Fig. \ref{figcomp} that in a fixed period of searching, the solutions obtained by NSJADE have the best convergence performance among the three MOEAs; and we give the credit to the powerful search engine of NSJADE: JADE.
Furthermore, since boundary points are one of the most important components in a PF, in Table \ref{tablecomp}, we record the mean values of the boundary points of the PF obtained in each run of 30 runs for these three algorithms. The nondominated boundary points are highlighted in gray background. According to the results in Table \ref{tablecomp}, it can be figured out that, 1) all the boundary points in the PF obtained by NSJADE are nondomindated; 2) only half of the boundary points in the PF obtained by NSCDE are nondominated; 3) and none of the boundary points in the PF obtained by NSGA-II are nondominated, which also means that NSJADE offers the best results when compared with NSGA-II and NSCDE.
\begin{table}[!htbp]
\scriptsize
\centering
\caption{Comparison of the mean values (mean$\pm$std) of the boundary points of the PF obtained in each run of 30 runs by NSGA-II, NSCDE, and NSJADE. The nondominated boundary points are highlighted in gray background.}\label{tablecomp}
\begin{tabular}{|c|c|c|c|c|}
\hline
& \multicolumn{4}{c|}{$\emph{S}_\emph{day}=-3$}\\\hline
& \multicolumn{2}{c|}{boundary point 1} & \multicolumn{2}{c|}{boundary point 2} \\\hline
& $f_1^{\textrm{eff}}$  & $f_2^{\textrm{eff}}$  & $f_1^{\textrm{eff}}$  & $f_2^{\textrm{eff}}$ \\\hline
NSGA-II & 8.3$\pm$4.4 & 12.4$\pm$4.2 & 23.2$\pm$5.3 & 0.2$\pm$0.5 \\\hline
NSCDE & \multicolumn{1}{>{\columncolor{mygray}}c}{0.7$\pm$0.7} & \multicolumn{1}{>{\columncolor{mygray}}c|}{14.5$\pm$4.2} & 17.3$\pm$3.2 & 0.0$\pm$0.0 \\\hline
NSJADE & \multicolumn{1}{>{\columncolor{mygray}}c}{0.9$\pm$1.3} & \multicolumn{1}{>{\columncolor{mygray}}c|}{10.8$\pm$2.6} & \multicolumn{1}{>{\columncolor{mygray}}c}{12.5$\pm$2.3} & \multicolumn{1}{>{\columncolor{mygray}}c|}{0.0$\pm$0.0} \\\hline
& \multicolumn{4}{c|}{$\emph{S}_\emph{day}=-7$}\\\hline
& \multicolumn{2}{c|}{boundary point 1} & \multicolumn{2}{c|}{boundary point 2} \\\hline
& $f_1^{\textrm{eff}}$  & $f_2^{\textrm{eff}}$  & $f_1^{\textrm{eff}}$  & $f_2^{\textrm{eff}}$ \\\hline
NSGA-II & 7.5$\pm$3.8 & 6.9$\pm$3.1 & 19.0$\pm$6.0 & 0.0$\pm$0.0 \\\hline
NSCDE & \multicolumn{1}{>{\columncolor{mygray}}c}{0.1$\pm$0.4} & \multicolumn{1}{>{\columncolor{mygray}}c|}{7.5$\pm$3.0} & 9.2$\pm$2.6 & 0.0$\pm$0.0 \\\hline
NSJADE & \multicolumn{1}{>{\columncolor{mygray}}c}{0.4$\pm$0.8} & \multicolumn{1}{>{\columncolor{mygray}}c|}{5.0$\pm$1.7} & \multicolumn{1}{>{\columncolor{mygray}}c}{6.2$\pm$1.9} & \multicolumn{1}{>{\columncolor{mygray}}c|}{0.0$\pm$0.0} \\\hline
& \multicolumn{4}{c|}{$\emph{S}_\emph{day}=-14$}\\\hline
& \multicolumn{2}{c|}{boundary point 1} & \multicolumn{2}{c|}{boundary point 2} \\\hline
& $f_1^{\textrm{eff}}$  & $f_2^{\textrm{eff}}$  & $f_1^{\textrm{eff}}$  & $f_2^{\textrm{eff}}$ \\\hline
NSGA-II & 9.8$\pm$5.0 & 14.1$\pm$5.8 & 31.2$\pm$16.1 & 1.2$\pm$0.9 \\\hline
NSCDE & \multicolumn{1}{>{\columncolor{mygray}}c}{0.2$\pm$0.5} & \multicolumn{1}{>{\columncolor{mygray}}c|}{12.9$\pm$3.4} & 21.6$\pm$9.7 & 0.0$\pm$0.2 \\\hline
NSJADE & \multicolumn{1}{>{\columncolor{mygray}}c}{1.5$\pm$1.7} & \multicolumn{1}{>{\columncolor{mygray}}c|}{9.1$\pm$3.2} & \multicolumn{1}{>{\columncolor{mygray}}c}{15.8$\pm$8.3} & \multicolumn{1}{>{\columncolor{mygray}}c|}{0.0$\pm$0.2} \\\hline
\end{tabular}
\end{table}

The experiments were carried out on a PC with $\textrm{Intel}^{\circledR}$ $\textrm{Core}\texttrademark$ i7 Processor 3.60GHz CPU and 8GB RAM. The processing time of each generation is 2 seconds, and it takes around 26.67 minutes each run for generating robust order schedules. It is worth mentioning that order scheduling is performed before the production, which can be regarded as an off-line scheduling. In addition, if high-performance computers and parallel computing are introduced to make the schedules in the factory, the scheduling time will further reduce. Meanwhile, intelligent order scheduling requires less manpower and fewer resources, which also saves the cost and increases the efficiency.

\section{Conclusion}
This paper has tackled the robust order scheduling problems in the fashion industry, which is a significant component of flexible and intelligent supply chain management.
Pre-production events in apparel manufacturing are taken into account for the first time, hence the order scheduling problem is modelled as a multi-objective optimization problem. In addition, in this paper, the daily production quantity of each order is assumed to be uncertain according to most real-world manufacturing environment of the fashion industry, which makes the problem into a robust multi-objective optimization problem. And NSJADE is utilized to search the robust order schedules.

A set of experiments have been carried out. The observations from the experiments show that pre-production events greatly influence the arrangements of the orders in the fashion industry. Moreover, it can be observed that the uncertainty in the daily production quantity of each order has a paramount impact on the order scheduling. The corresponding robust PFs are also provided under various settings of parameters including the uncertainty factor $\beta$ and the number of the neighbouring solutions $H$. It is found that robust order schedules can be shifted less often after the production starts than non-robust ones, which saves labor cost and enhances the production efficiency. Meanwhile, with the help of robust order schedules, planners can pay close attention to the unfinished pre-production events as early as possible, negotiate earlier with the customers who place the orders about the delay in delivery, or arrange operators to work extra hours for these orders.

\ifCLASSOPTIONcaptionsoff
  \newpage
\fi

{\footnotesize\bibliography{ref}
\bibliographystyle{ieeetr}}
\clearpage

\section*{\Large{\textbf{Supplementary file}}}

\vspace{1cm}

\section*{Section Captions}
\begin{itemize}
\item[$\bullet$] \textbf{Section S.I} The Other Operations
\item[$\bullet$] \textbf{Section S.II} Experimental Setup
\item[$\bullet$] \textbf{Section S.III} The Effect of $\xi$ on Search Performance
\end{itemize}

\section*{Table Captions}
\begin{itemize}
\item[$\bullet$] \textbf{Table \ref{podets}} The details of the 20 production orders collect from Fast React.
\item[$\bullet$] \textbf{Table \ref{pldets}} The details of the 6 production lines collected from Fast React.
\item[$\bullet$] \textbf{Table \ref{odsb}} The details of the order assignments on 6 production lines in schedule SB.
\end{itemize}

%\newpage

\section*{Figure Captions}
\begin{itemize}
\item[$\bullet$] \textbf{Fig. \ref{figab}} An illustration of robust solutions from decision space to objective space. Solution A is more robust to perturbations in variable than solution B when two objectives $f_1$ and $f_2$ are optimized.
%\item[$\bullet$] \textbf{Fig. \ref{flchart}} Flowchart of the proposed NSJADE.
\item[$\bullet$] \textbf{Fig. \ref{figlcs}} The learning curves of producing different products. (a) Leaning curve of producing skirts and pants; (b) Learning curve of producing blouses; (c) Learning curve of producing jackets.
\item[$\bullet$] \textbf{Fig. \ref{plotpf}} The Pareto fronts sorted out from the solutions obtained after 30 runs by NSJADE when $\xi=[5, 10, 15]$.
\end{itemize}

\clearpage

\section*{S.I The Other Operations}
\subsection{Fast Nondominated Sorting and Crowding-Distance Assignment}
In our proposed NSJADE, we keep the fast nondominated sorting and crowding-distance assignment, which are two effective mechanisms presented in NSGA-II [20]. After evaluating the population, a fast nondominated sorting approach is employed to sort the population into different nondomination levels with a lower computational complexity compared to the traditional approach. For each individual, the average distance of two individuals on either side of this individual is calculated along each of the objectives, and the distance is called crowding distance. When two individuals are in the same nondomination level, the individual with a larger value of crowding distance is preferred. For more details of these two mechanisms, one can refer to [20].

\subsection{Mutation and Crossover}
In the developed NSJADE, adaptive differential evolution (JADE) is elected as the search engine instead of the non-adaptive genetic algorithm for promoting both exploration and exploitation abilities of the population. Therefore, the mutation and crossover strategies of JADE are adopted in NSJADE. One can refer to [21] for the details of the mutation and crossover strategies.

\subsection{Selection}
In a single evolution, after each individual in the parent population goes through the mutation and crossover, a new generation needs to be selected from the combined parent and the offspring population. It is assumed that the parent population contains \emph{NP} individuals; therefore, after a single evolution, there are \emph{2NP} individuals (each parent generates one offspring) in the candidate pool. Sort the population into different nondomination levels by means of the fast nondominated sorting approach, and calculate the crowding distance of each individual. The individual with the lower nondomination level and the larger crowding distance is preferred. By following this principle, \emph{NP} individuals are chosen from the candidate pool as the population of a new generation.

\section*{S.II Experimental Setup}
In this section, we introduce the experimental setup in detail, including the information of the test data, the production orders, and the production lines.

\subsection*{A. Test Data Information}
The test data used for the following experiments are gathered from a business software called Fast React [17]. Fast React is a production planning software for the fashion industry, which considers practical factors in real-world production, like pre-production events, learning effect, etc. And Fast React has been used by global fashion brands, high street retailers and prominent worldwide manufacturers involved in the real business ranging from carpet, cloth and lace weavers to shoe manufacturers, clothing companies and so on. Fast React provides a lot of industrial data collected from their customers, and the data include different types of orders and production lines as well as the information of the orders' pre-production events. Therefore, these data are utilized as the test data in our research.

Although Fast React simulates the real production in the fashion industry, it does not consider various uncertainties that exist in the real-world manufacturing. In addition, planners are required to manually place the orders on the production lines in Fast React, which is not intelligent. In this research, uncertainties in the daily production quantity are taken into consideration; and robust order schedules can be intelligently obtained by using our proposed NSJADE.

\subsection*{B. Production Order Information}
Each production order has five attributes: product type, quantity, present conservative starting date, due date, and standard minutes per piece of this order. The detailed descriptions of each attribute can be found in Section II-A.
There are total four categories of orders: skirts, blouses, pants and jackets. The learning curves of producing these four types of products are collected from Fast React [17] and provided in Fig. \ref{figlcs} in the supplementary file. A total of 20 orders are collected for the following experiments. The details of these orders are given in Table \ref{podets} in the supplementary file.

\subsection*{C. Production Line Information}
As in Fast React [17], there are total 6 production lines considered in the experiments. These production lines are product-specific lines, which means that the line's efficiency is lower than its peak value when there is a mismatch of product to production line. The details of the production lines are listed in Table \ref{pldets} in the supplementary file.

\section*{S.III The Effect of $\xi$ on Search Performance}
The parameter $\xi$ is to balance the search accuracy and the computational complexity of NSJADE. A large $\xi$ is apt to enhance the search accuracy and result in a heavy computational complexity; a small $\xi$ is able to reduce the computational burdens and may bring about unsatisfactory search performance. Therefore, an appropriate tuning of $\xi$ must be carried out. In this subsection, a set of tests are conducted to select a proper $\xi$ for the order scheduling problem studied in this paper. We make the order schedule 3 days before the production begins, i.e., $\emph{S}_\emph{day}=-3$, and uncertainty is not considered. $\xi$ is set as $[5, 10, 15]$. The Pareto fronts (PFs) obtained by NSJADE with three different values of $\xi$ are provided in Fig. \ref{plotpf}.
It can be observed that the PFs overlap completely when $\xi=10$ and $\xi=15$, and the solutions on these two PFs dominate the solutions on the PF when $\xi=5$. As a result, in order to achieve a balance between search accuracy and computational complexity, we select $\xi=10$ for NSJADE in this paper.

\clearpage

\setcounter{table}{0}
\renewcommand\thetable{S.\arabic{table}}
\begin{table*}[!htbp]
\footnotesize
\centering
\caption{The details of the 20 production orders collect from Fast React.}\label{podets}
\begin{tabular}{|c|c|c|c|c|c|}
\hline
\tabincell{c}{\textbf{Production} \\ \textbf{Order No.}} & \tabincell{c}{\textbf{Product} \\ \textbf{Type}} & \textbf{Quantity} & \tabincell{c}{\textbf{Present Conservative} \\ \textbf{Starting Date} ($\emph{\textbf{S}}_\emph{\textbf{day}}\textbf{=-3, -7, -14}$)} & \textbf{Due Date} & \tabincell{c}{\textbf{Standard Minutes} \\ \textbf{Per Piece}} \\\hline
1  & Skirts & 870 & 0, 0, 6 & 10 & 14.20  \\\hline
2  & Skirts & 700 & 0, 0, 6 & 7 & 18.20   \\\hline
3  & Blouses & 800 & 0, 0, 6 & 11 & 18.20  \\\hline
4  & Skirts & 500 & 0, 3, 0 & 9 & 18.20  \\\hline
5  & Skirts & 1000 & 0, 0, 0 & 11 & 16.70  \\\hline
6  & Skirts & 1000 & 7, 3, 0 & 10 & 16.70  \\\hline
7  & Blouses & 800 & 0, 0, 6 & 7 & 32.20  \\\hline
8  & Jackets & 850 & 12, 8, 1 & 15 & 54.60  \\\hline
9  & Skirts & 800 & 0, 0, 6 & 10 & 16.70  \\\hline
10  & Pants & 780 & 12, 8, 1 & 15 & 34.00  \\\hline
11  & Blouses & 1000 & 0, 0, 11 & 15 & 15.00  \\\hline
12  & Jackets & 1000 & 0, 0, 6 & 8 & 53.78  \\\hline
13  & Skirts & 400 & 22, 18, 11 & 24 & 26.50  \\\hline
14  & Blouses & 2000 & 0, 13, 6 & 15 & 12.60  \\\hline
15  & Blouses & 1000 & 4, 0, 0 & 11 & 12.60  \\\hline
16  & Jackets & 500 & 12, 8, 1 & 18 & 44.10  \\\hline
17  & Skirts & 800 & 0, 0, 11 & 15 & 20.55  \\\hline
18  & Skirts & 800 & 0, 0, 0 & 19 & 20.55  \\\hline
19  & Jackets & 700 & 0, 18, 11 & 20 & 44.10  \\\hline
20  & Blouses & 3000 & 17, 13, 6 & 19 & 12.60  \\\hline
\end{tabular}
\end{table*}

\renewcommand\thetable{S.\arabic{table}}
\begin{table*}[!htbp]
\footnotesize
\centering
\caption{The details of the 6 production lines collected from Fast React.}\label{pldets}
\begin{tabular}{|c|c|c|c|c|}
\hline
\tabincell{c}{\textbf{Production} \\ \textbf{Line No.}} & \tabincell{c}{\textbf{Efficiency for} \\ \textbf{Skirts/Pants (\%)}} & \tabincell{c}{\textbf{Efficiency for} \\ \textbf{Blouses (\%)}} & \tabincell{c}{\textbf{Efficiency for} \\ \textbf{Jackets (\%)}} & \tabincell{c}{\textbf{Capacity} \\ \textbf{(mins/day)}} \\\hline
1  & 100 & 80 & 80 & 6720  \\\hline
2  & 100 & 80 & 80 & 6720  \\\hline
3  & 80 & 100 & 80 & 6240  \\\hline
4  & 80 & 100 & 80 & 6240  \\\hline
5  & 80 & 80 & 100 & 6720  \\\hline
6  & 80 & 80 & 100 & 6720  \\\hline
\end{tabular}
\end{table*}

\renewcommand\thetable{S.\arabic{table}}
\begin{table*}[!htbp]
\footnotesize
\centering
\caption{The details of the order assignments on 6 production lines in schedule SB.}\label{odsb}
\begin{tabular}{|c|c|}
\hline
\tabincell{c}{\textbf{Production} \\ \textbf{Line No.}} & \textbf{Order Assignments} \\\hline
 1 & \tabincell{c}{$O_5(800)$, $O_{18}(640)$, $O_{2}(560)$, $O_1(870)$, $O_9(800)$,\\ $O_{17}(800)$}   \\\hline
 2 & \tabincell{c}{$O_6(400)$, $O_{10}(624)$, $O_{2}(140)$, $O_3(320)$, $O_{14}(800)$,\\ $O_{11}(1000)$, $O_{13}(400)$}  \\\hline
 3 & \tabincell{c}{$O_5(200)$, $O_{6}(600)$, $O_{15}(400)$, $O_7(320)$, $O_{14}(1200)$,\\ $O_{20}(1200)$}  \\\hline
 4 & \tabincell{c}{$O_{18}(160)$, $O_{10}(156)$, $O_{15}(600)$, $O_{7}(480)$, $O_{3}(480)$,\\ $O_{20}(1800)$}  \\\hline
 5 & $O_{4}(300)$, $O_{16}(200)$, $O_{12}(600)$, $O_{8}(170)$  \\\hline
 6 & $O_{4}(200)$, $O_{16}(300)$, $O_{12}(400)$, $O_{8}(680)$, $O_{19}(700)$  \\\hline
\end{tabular}
\end{table*}

\clearpage

\setcounter{figure}{0}
\renewcommand\thefigure{S.\arabic{figure}}
\begin{figure*}
\centering
\subfigure[]{
\begin{minipage}[b]{0.35\textwidth}
\includegraphics[width=1\textwidth]{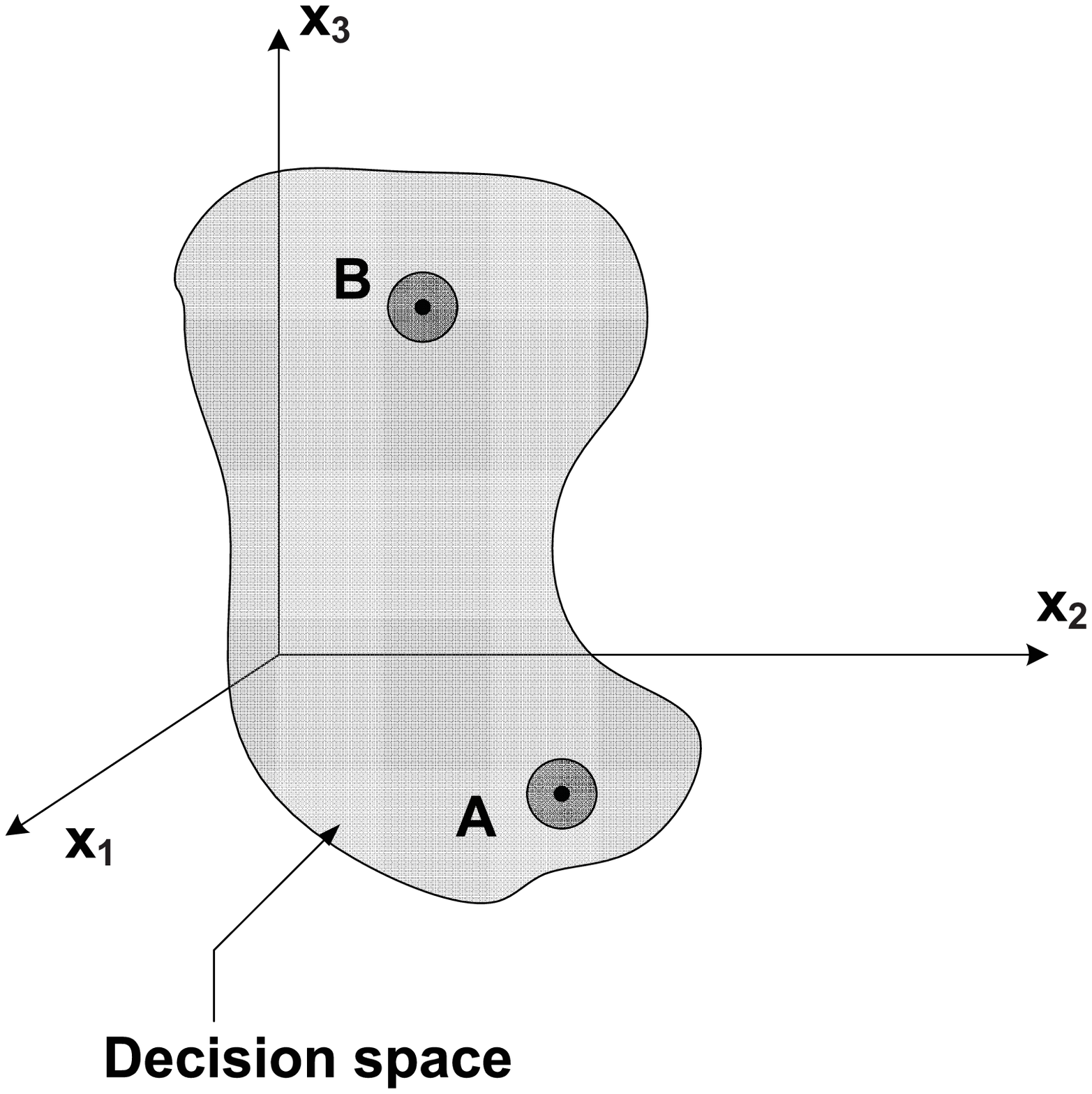}
\end{minipage}
}
\subfigure[]{
\begin{minipage}[b]{0.35\textwidth}
\includegraphics[width=0.95\textwidth]{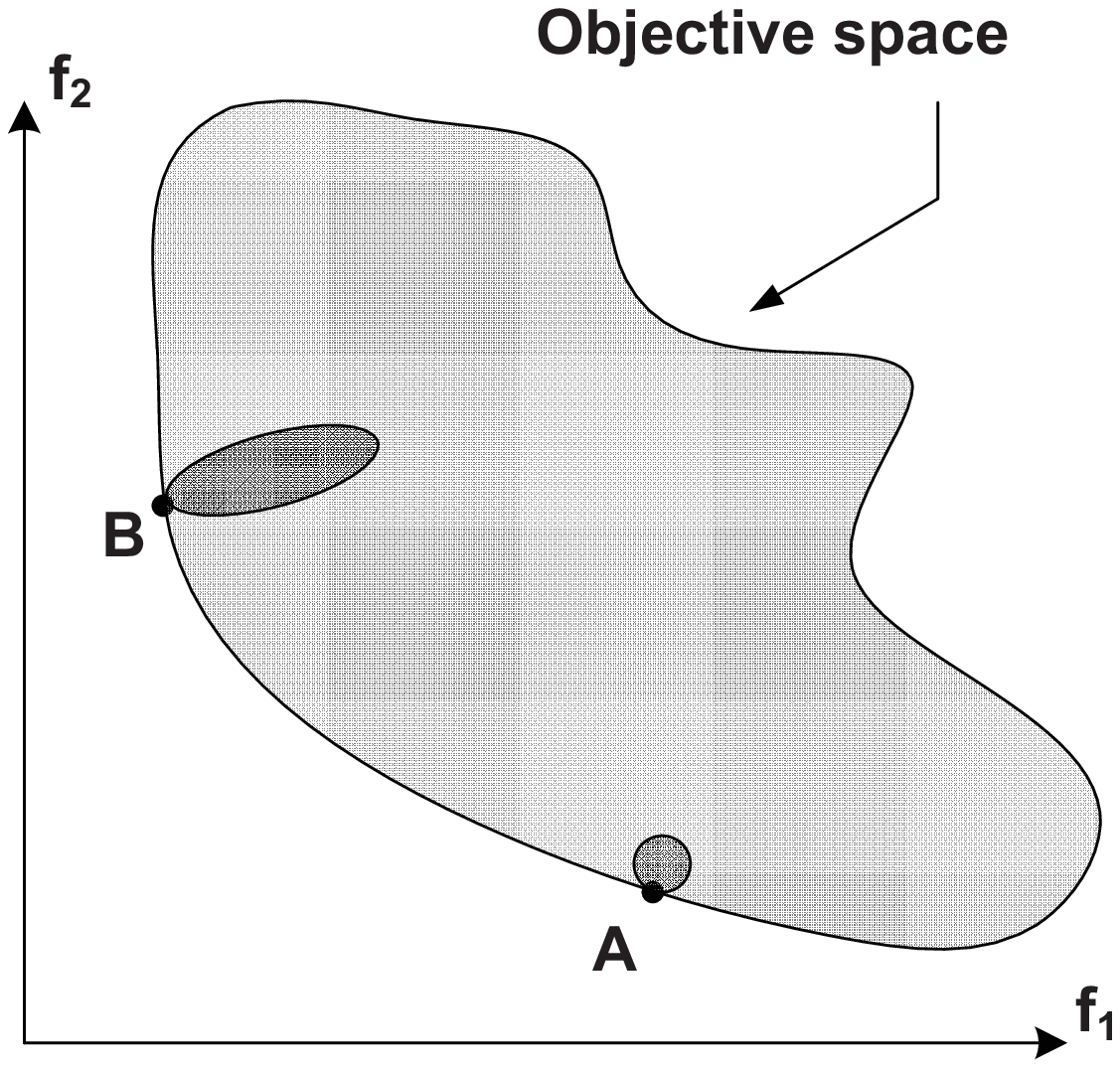}
\end{minipage}
}
\caption{An illustration of robust solutions from decision space to objective space. Solution A is more robust to perturbations in variable than solution B when two objectives $f_1$ and $f_2$ are optimized.} \label{figab}
\end{figure*}

\renewcommand\thefigure{S.\arabic{figure}}
\begin{figure*}
\centering
\subfigure[]{
\begin{minipage}[b]{0.35\textwidth}
\includegraphics[width=0.95\textwidth]{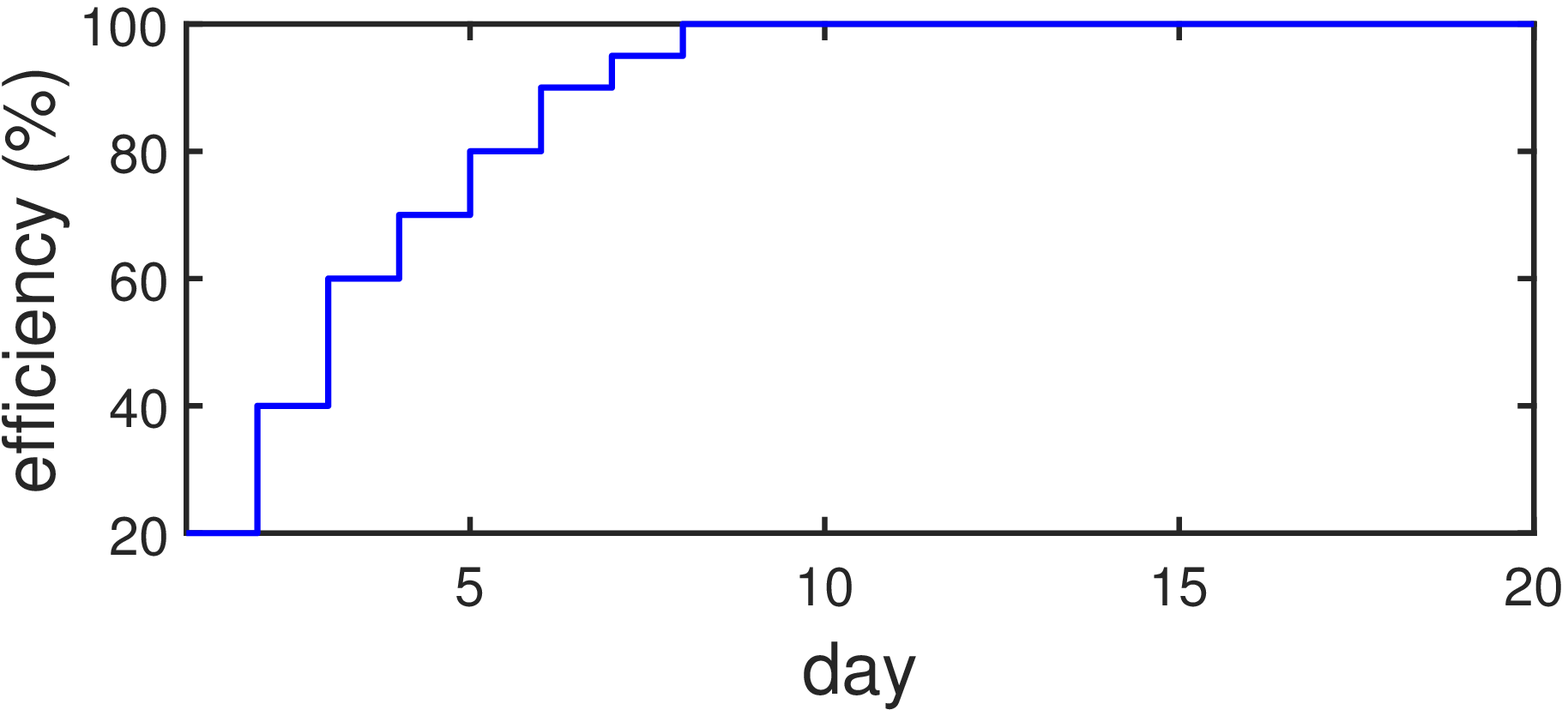}
\end{minipage}
}
\subfigure[]{
\begin{minipage}[b]{0.35\textwidth}
\includegraphics[width=0.95\textwidth]{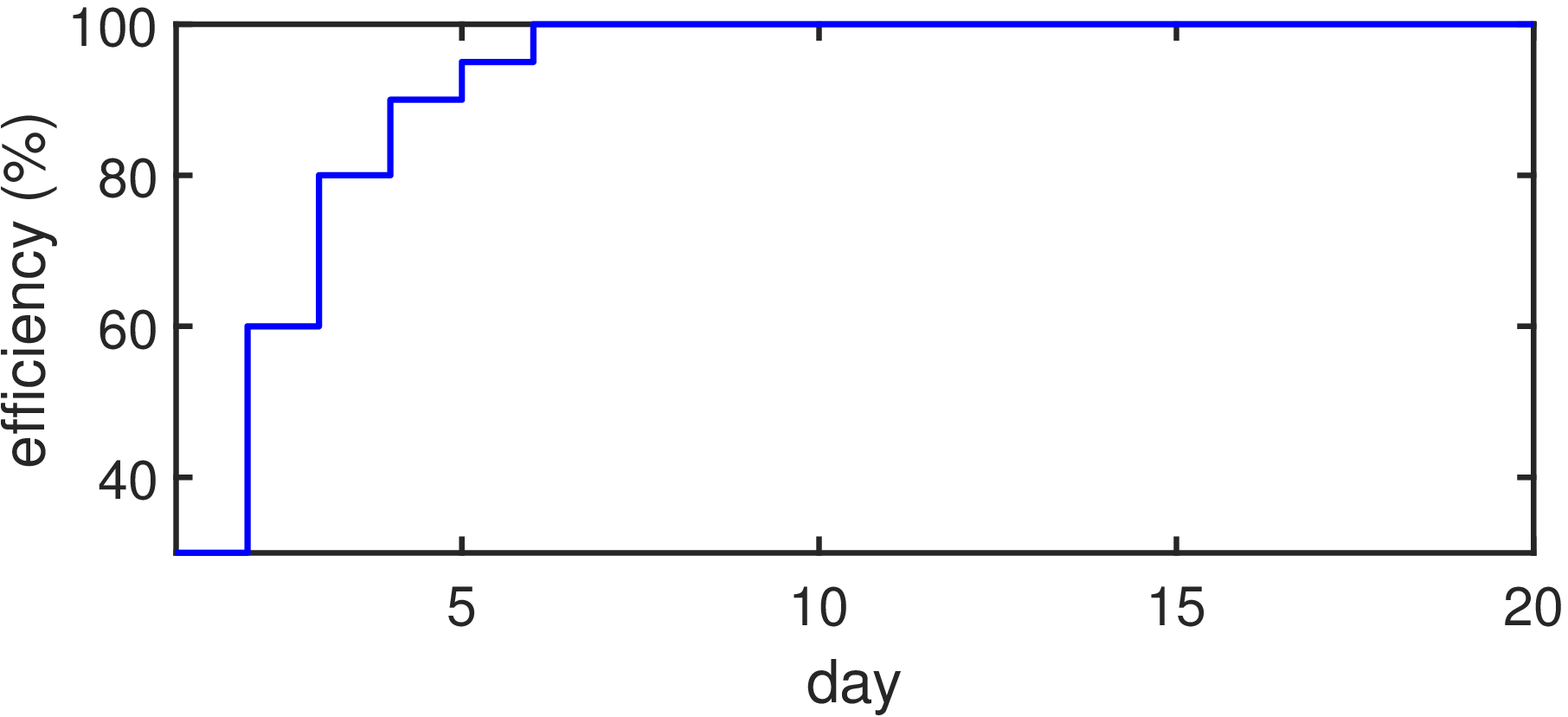}
\end{minipage}
}
\subfigure[]{
\begin{minipage}[b]{0.35\textwidth}
\includegraphics[width=0.95\textwidth]{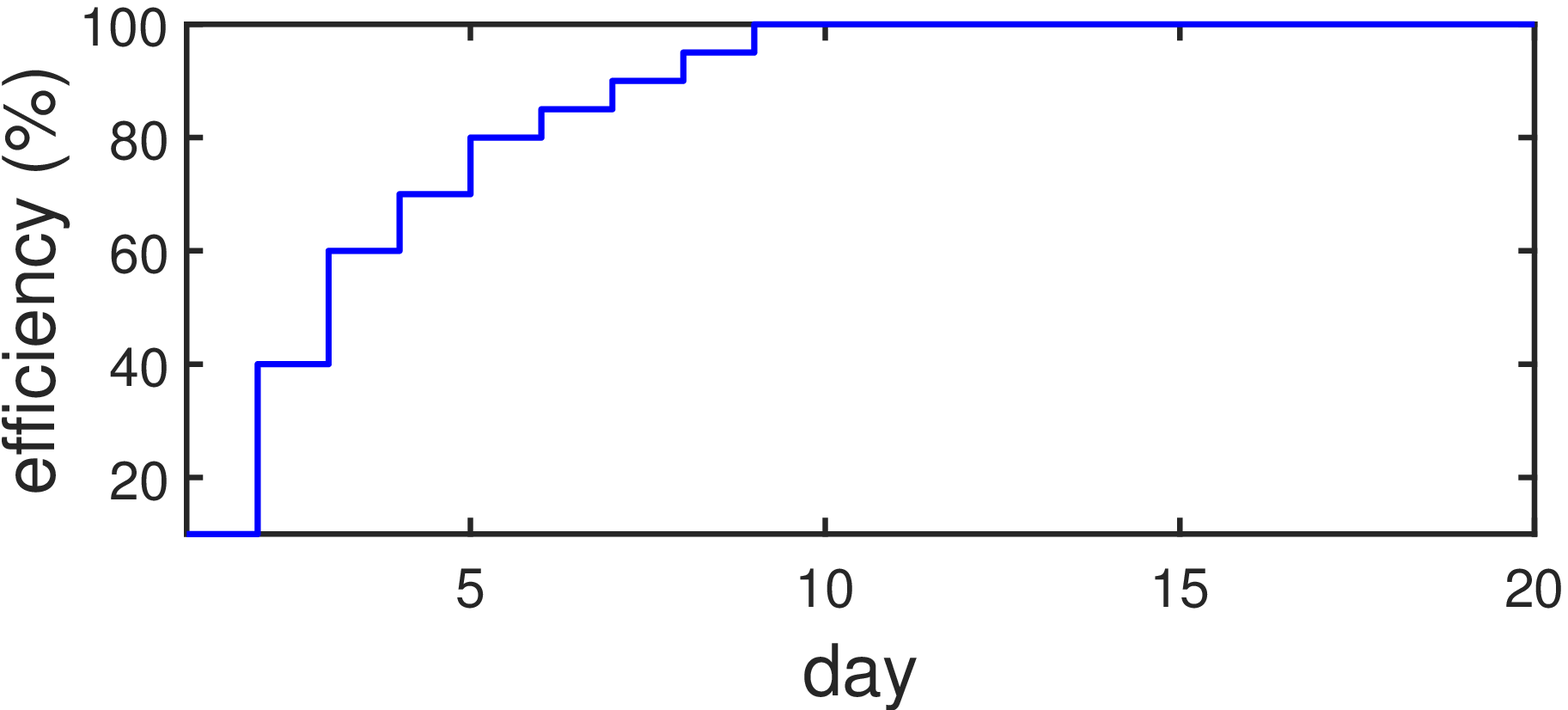}
\end{minipage}
}
\caption{The learning curves of producing different products. (a) Leaning curve of producing skirts and pants; (b) Learning curve of producing blouses; (c) Learning curve of producing jackets.} \label{figlcs}
\end{figure*}

\renewcommand\thefigure{S.\arabic{figure}}
\begin{figure*}[!htbp]
\begin{minipage}[t]{1\linewidth}
\centering
\includegraphics[width=6cm]{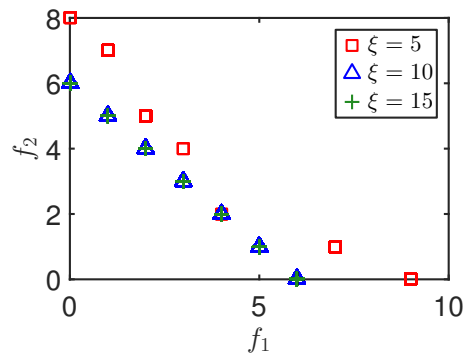}
\caption{The Pareto fronts sorted out from the solutions obtained after 30 runs by NSJADE when $\xi=[5, 10, 15]$.} \label{plotpf}
\end{minipage}
\end{figure*}

\end{document}